\documentclass[letterpaper, 10 pt, journal, twoside]{IEEEtran}

\IEEEoverridecommandlockouts                              

\usepackage{Sweave}
\usepackage[noadjust]{cite}
\usepackage{amsmath,amssymb,amsfonts}
\usepackage{algorithmic}
\usepackage{graphicx}
\usepackage{textcomp}
\usepackage{multirow}
\usepackage[dvipsnames]{xcolor} 
\usepackage[inline]{enumitem}
\usepackage{comment}
\usepackage{subcaption}
\usepackage{siunitx}
\usepackage{flushend}
\def\BibTeX{{\rm B\kern-.05em{\sc i\kern-.025em b}\kern-.08em
    T\kern-.1667em\lower.7ex\hbox{E}\kern-.125emX}}

\usepackage[symbol]{footmisc}

\begin{document}
\title{Bio-LSTM: A Biomechanically Inspired Recurrent Neural Network for 3D Pedestrian Pose \\ and Gait Prediction}

\markboth{IEEE Robotics and Automation Letters. Preprint Version. Accepted
January 2019}
{Du \MakeLowercase{\textit{et al.}}: Bio-LSTM:  A Biomechanically Inspired RNN for 3D Pedestrian Pose and Gait Prediction} 

\author{Xiaoxiao Du$^{1}$, Ram Vasudevan$^{2}$, and Matthew Johnson-Roberson$^{1}$
\thanks{This work was supported by a grant from Ford Motor Company via the Ford-UM Alliance under award N022884.} 
\thanks{$^{1}$X. Du and M. Johnson-Roberson are with Department of Naval Architecture and Marine Engineering, University of Michigan, Ann Arbor, MI 48109 USA
        {\tt\footnotesize xiaodu@umich.edu; mattjr@umich.edu}}%
\thanks{$^{2} $R. Vasudevan is with the Department of Mechanical Engineering, University of Michigan, Ann Arbor, MI 48109 USA
        {\tt\footnotesize ramv@umich.edu}}%

}

\maketitle



\begin{abstract}
In applications such as autonomous driving, it is important to understand, infer, and anticipate the intention and future behavior of pedestrians. This ability allows vehicles to avoid collisions and improve ride safety and quality. This paper proposes a biomechanically inspired recurrent neural network (Bio-LSTM) that can predict the location and 3D articulated body pose of pedestrians in a global coordinate frame, given 3D poses and locations estimated in prior frames with inaccuracy. The proposed network is able to predict poses and global locations for multiple pedestrians simultaneously, for pedestrians up to 45 meters from the cameras (urban intersection scale). The outputs of the proposed network are full-body 3D meshes represented in Skinned Multi-Person Linear (SMPL) model parameters. The proposed approach relies on  a novel objective function that incorporates the periodicity of human walking (gait), the mirror symmetry of the human body, and the change of ground reaction forces in a human gait cycle.  This paper presents prediction results on the PedX dataset, a large-scale, in-the-wild data set collected at real urban intersections with heavy pedestrian traffic. Results show that the proposed network can successfully learn the characteristics of pedestrian gait and produce accurate and consistent 3D pose predictions.
\end{abstract}


\section{Introduction}
\label{sec:intro}
\IEEEPARstart{I}{magine} that an autonomous vehicle is driving towards a crowded urban intersection. It is important to identify moving pedestrians and anticipate where a pedestrian, or a group of pedestrians, may be in a few seconds to decide whether and when to brake.  Imagine also that a robot is serving as a tour guide in a museum \cite{thrun1999minerva} or in a shopping mall packed with pedestrians \cite{luo2018porca}. It is essential for the robot to recognize the orientation and location of persons around to provide better guidance and avoid running into pedestrians.  In these scenarios, accurate pedestrian pose and location prediction has a huge impact in facilitating more effective human-robot/vehicle interaction and collision avoidance.

Human pose estimation has been heavily studied in the literature \cite{li20143d, park20163d, toshev2014deeppose, simo2013joint, guler2018densepose, zimmermann20183d}. However, prior work has primarily focused on estimating the joint locations of a human skeleton model from a single, static RGB image in the current frame and does not address the pose-prediction problems for future frames. More recently, researchers have begun investigating the prediction (forecasting and anticipation) of human body pose given a video sequence \cite{chao2017forecasting, toyer2017human, fragkiadaki2015recurrent, walker2017pose, pavllo2018quaternet, jain2016structural, zanfir2018monocular, martinez2017human}. Most of this work focuses on a skeleton-based representation for joint locations. Moreover, some studies such as  \cite{toyer2017human, fragkiadaki2015recurrent} are limited to predicting the 2D pose of a single human subject, usually centered in a video frame. On the other hand, deep learning techniques, especially recurrent neural networks, have proven to be effective in predicting future frames in natural video sequences \cite{lotter2016deep, villegas2017decomposing}. However, these approaches focus on pixel-level prediction on images and do not specifically work with human pose representations (skeleton or mesh). 

This paper focuses on two novel aspects of the problem: predicting a full-body 3D mesh and doing so for multiple humans simultaneously. Furthermore, we attempt to constrain the problem using the well-studied biomechanics of human walking while using the contextual information within urban-intersection environments. Note that in some of the literature, the terms ``pose prediction'' and ``pose estimation'' are used interchangeably, both referring to the task of estimating a pose (usually skeleton-based joint locations) from a single image (the current frame) \cite{tan2017indirect, pavlakos2018learning}. In this paper, we use the term ``prediction'' to refer to the specific task of predicting/forecasting 3D pedestrian pose and location in future frames in a sequence, assuming the 3D poses were already estimated in a prior frame. The estimation of the initial 3D pose model is outside of the scope of this paper, but is described in depth in Kim et al.~\cite{kim2018pedx}

We propose bio-LSTM, a biomechanically inspired recurrent neural network to solve this task. The proposed network takes previously estimated pose parameters in past frames as input and outputs a full-body 3D mesh of a pedestrian pose, localized in a global coordinate system in  metric space at future timesteps. Our network can predict multiple pedestrians in each frame at real intersection scales (up to 45 meters), and the mesh representation contains richer information about the body shape and scale that traditional skeletal representations lack \cite{bogo2016keep}.  The proposed network is based on the long short-term memory (LSTM) network \cite{hochreiter1997long} with inspiration from the biomechanics of human gait, such as the bilateral/mirror symmetry of the human body \cite{troje2005person}, the periodicity of human walking (gait) \cite{vaughan1999dynamics}, and the change of ground reaction force in a human gait cycle \cite{kong2008smooth, winter1995human}.

 We present experimental results of our proposed network on the PedX dataset \cite{kim2018pedx}, a large-scale, in-the-wild dataset collected at real urban intersections with heavy pedestrian traffic in the city of Ann Arbor, Michigan, USA. In addition to the PedX intersection dataset, we also collected and annotated an evaluation dataset in a controlled outdoor environment with a motion capture (mocap) system. We compare our prediction to both the 3D labels generated by a novel optimization procedure~\cite{kim2018pedx} and  the mocap ground truth to verify the accuracy of our method. Results show successful and accurate body pose prediction for both next-frame and multiple timesteps.

\textcolor{black}{The contributions of this paper include: \begin{enumerate*}
\item full-body 3D mesh prediction in addition to skeleton-based joint locations in global coordinate frame and in metric space;
\item a novel biomechanics-based loss function in the LSTM network to ensure realistic and naturalistic pose prediction; and
\item in-the-wild gait and pose prediction for multiple pedestrians given noisy urban intersection data. 
\end{enumerate*}
 We envision our work having applications in the development of legged robots, rehabilitation, and robot-assisted physiotherapy, in addition to our original motivation in the autonomous driving and human-robot interaction contexts. We present longer-term prediction results, which also enables evasive maneuvers and path planning using the prediction information as well as semantic interpretation of the pedestrian's actions in the future. }

This paper is organized as follows: Section~\ref{sec:intro} introduces the problem of 3D human forecasting and motivates our work. Section~\ref{sec:relatedwork} describes related work in sequence prediction and introduces the SMPL model~\cite{loper2015smpl}, a parametric body-shape model that we use to represent the 3D human pose. We also describe related work in gait analysis, where we drew inspiration for our network formulation. Section~\ref{sec:method} describes our proposed network and bio-inspired loss function. Section~\ref{sec:setup} describes the PedX dataset and the experimental setup. Section~\ref{sec:results} presents our prediction results on both next-frame and multiple frame forecasts. Section~\ref{sec:conclusion} presents our conclusions and future work.

\section{Related Work}
\label{sec:relatedwork}
In this section, we first describe related works on video sequence prediction. Then, we describe the SMPL model that we use to represent 3D human pose. We also describe the related works in gait biomechanics that inspired our method.


\subsection{Sequence Prediction}
Recurrent Neural Networks (RNN) have shown effective results in learning temporal dynamics in a sequence \cite{lipton2015critical}. The LSTM network \cite{hochreiter1997long}, in particular, has been widely used in the literature for sequence prediction due to its ability to learn long-term dependencies \cite{sun2017lattice, park2018sequence, liu2017dap3d, sun20183dof}. Recently, the LSTM networks have been applied to predicting future image-based frames in natural video sequences, such as PredNet \cite{lotter2016deep} and MCnet \cite{villegas2017decomposing}. However, these studies mainly focus on video image sequences and usually use convolution operations to take advantage of the pixel spatial layout in the image. 

\begin{figure}[h!]
\centering
\begin{subfigure}{0.12\textwidth}
\includegraphics[scale=0.17, trim={0 10mm 0 0},clip]{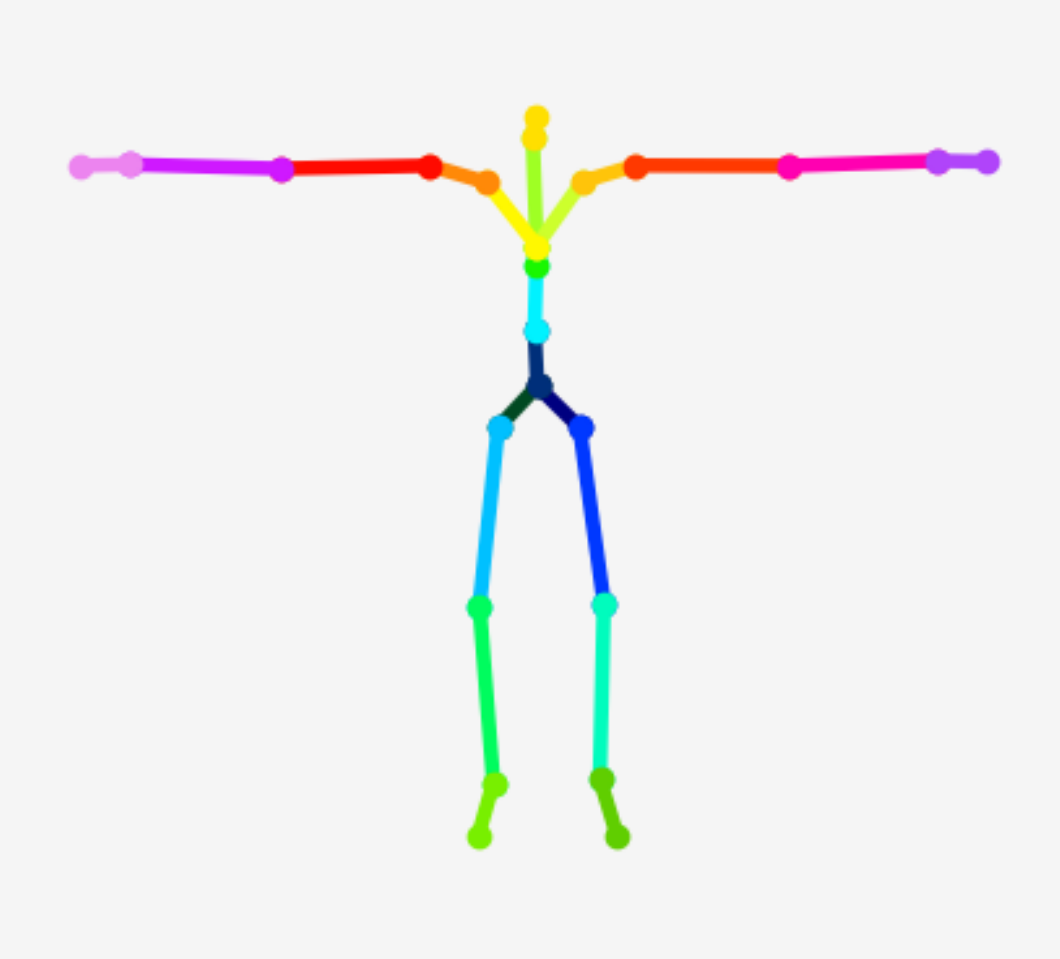}
\caption{ }
\label{fig:skeleton_restpose_big}
\end{subfigure}
\begin{subfigure}{0.11\textwidth}
\includegraphics[scale=0.12, trim={7mm 2mm 5mm 5mm},clip]{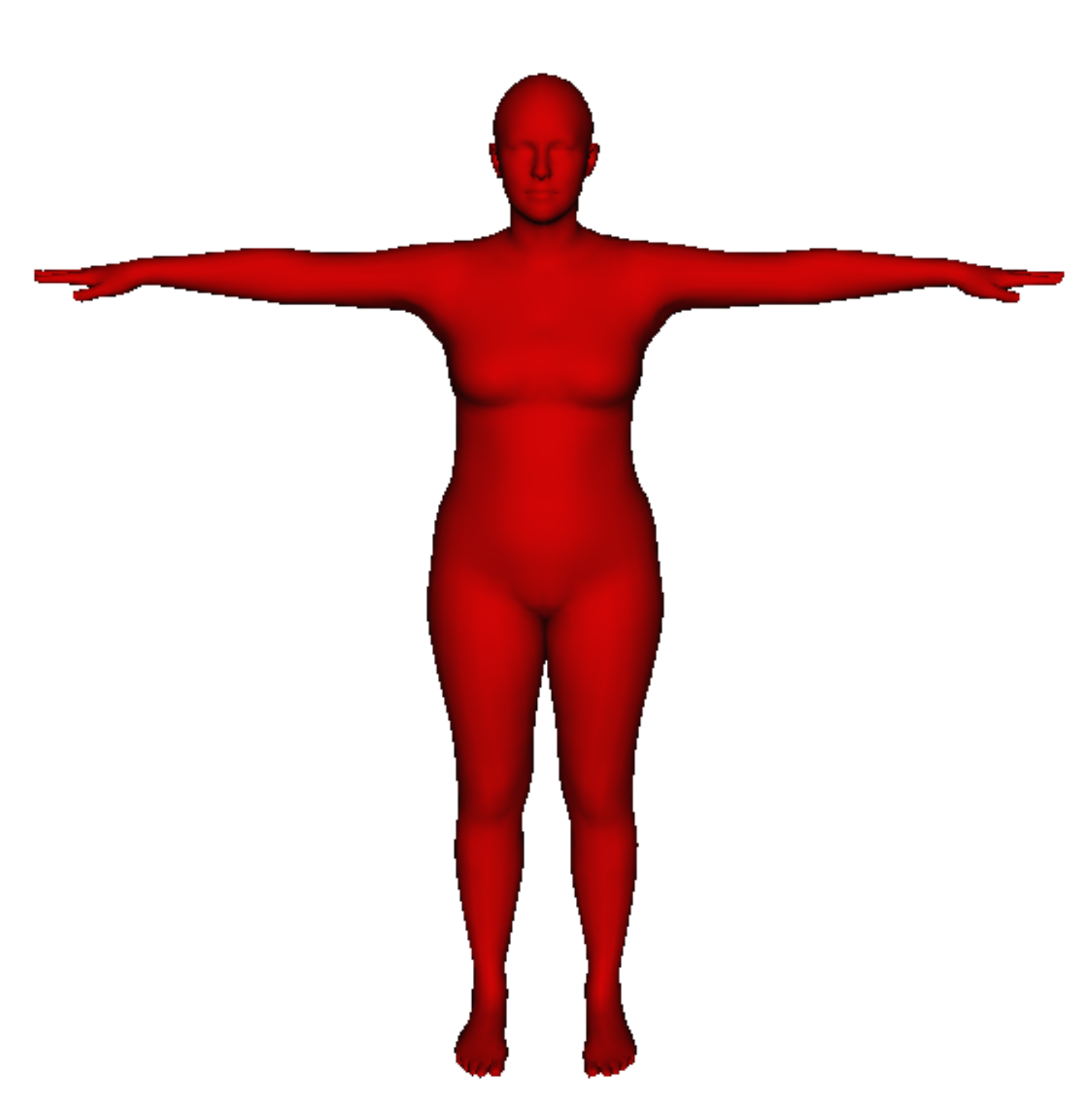}
\caption{ }
\label{fig:skeleton_restpose_true}
\end{subfigure}
\begin{subfigure}{0.12\textwidth}
\includegraphics[scale=0.09, trim={0mm 0 0mm 0},clip]{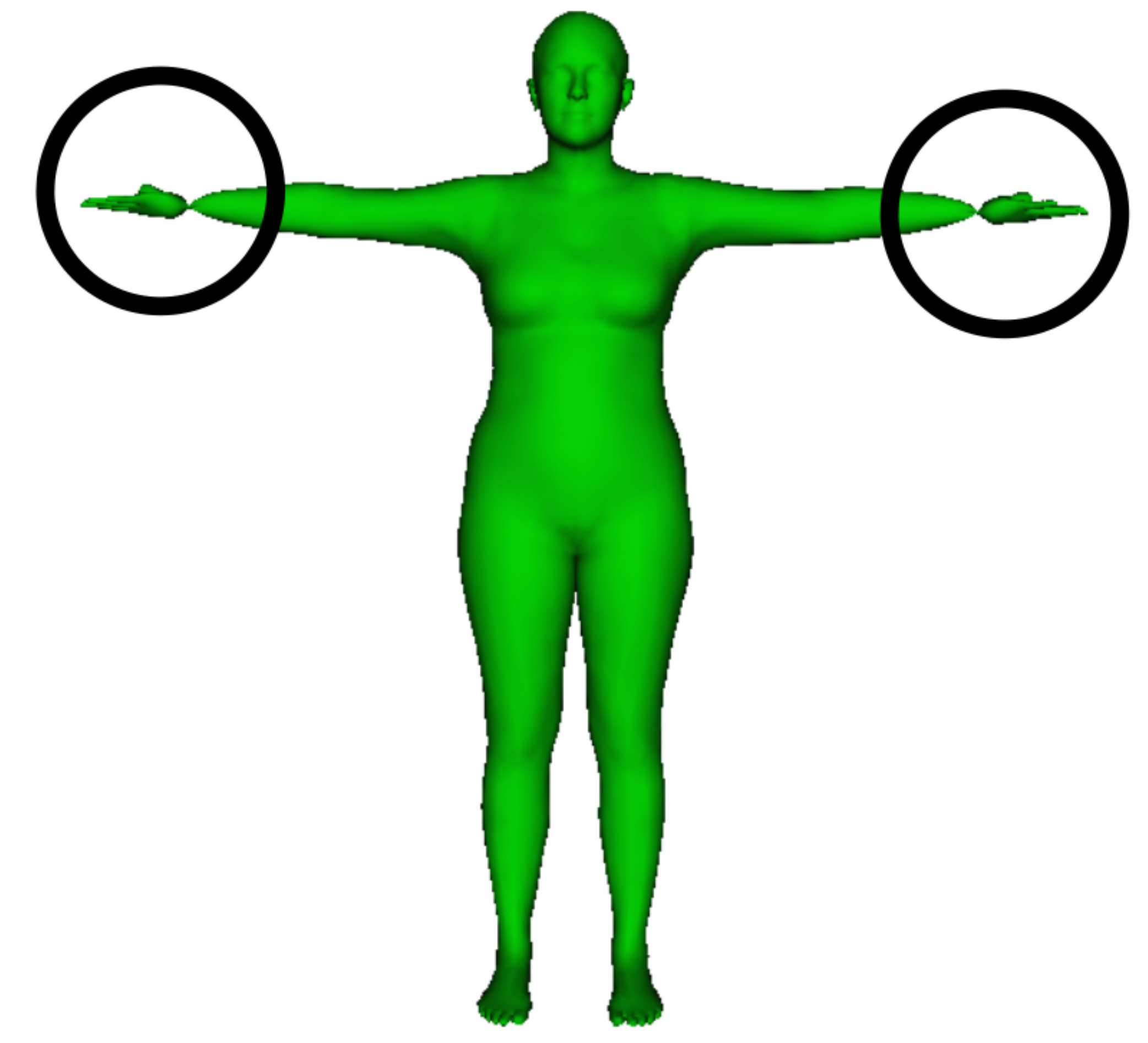}
\caption{ }
\label{fig:skeleton_restpose_wrong}
\end{subfigure}
\begin{subfigure}{0.12\textwidth}
\includegraphics[scale=0.11, trim={0 0 0 0},clip]{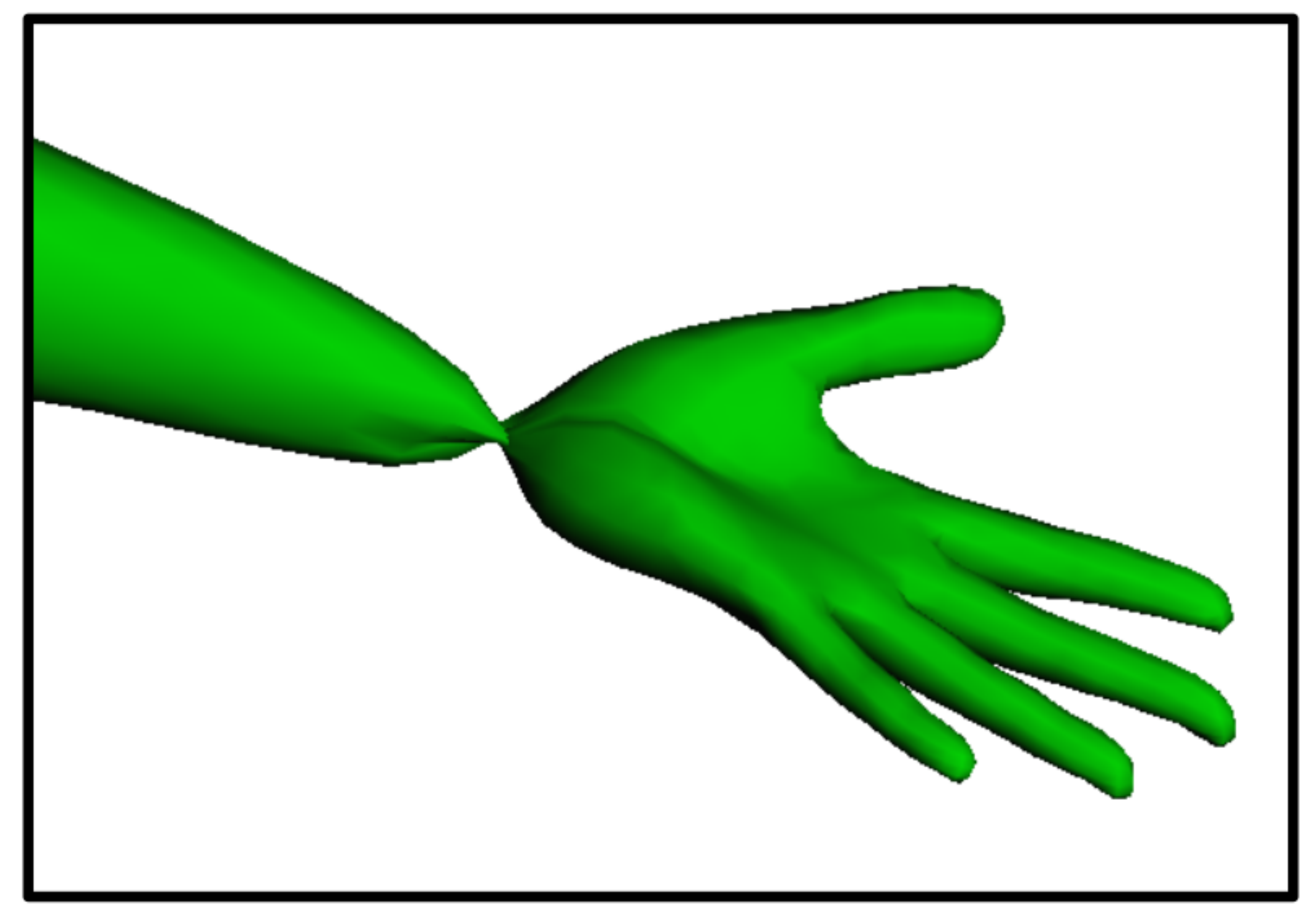}
\caption{ }
\label{fig:hand}
\end{subfigure}
\caption{An illustration of a human pose skeleton and full-body mesh. (a) A rest pose (T-pose) skeleton. (b) A SMPL \cite{loper2015smpl} full-body mesh at the rest pose. (c) Another SMPL full-body mesh with the same skeleton joint location as (a), but with biologically unfeasible wrist rotations (marked in circles). (d) The zoomed-in view of the biologically unfeasible wrist-joint in (c). The rotation on the wrist has turned $\pi$ degrees, but the wrist joint location remains the same.}
\label{fig:skeletonmesh}
\end{figure}

For the specific task of human pose prediction,  previous research has investigated predicting joint locations in future frames given past video sequences \cite{chao2017forecasting, toyer2017human, fragkiadaki2015recurrent, walker2017pose}. However, in most of these studies, the human pose is represented simply by joint locations in a skeleton and visualized by overlaying the skeleton on the 2D image.  Moreover, Toyer et al. \cite{toyer2017human} and Fragkiadak et al. \cite{fragkiadaki2015recurrent} are limited to predicting 2D pose for a single human subject centered in a video sequence. However, these assumptions do not always hold. For videos collected at a crowded urban intersection, there are multiple pedestrians moving simultaneously, and some pedestrians can be quite far from the camera. Additionally, skeleton-based joint locations may not always accurately represent the full human-body pose. For example, Figure~\ref{fig:skeleton_restpose_true} and Figure~\ref{fig:skeleton_restpose_wrong} both have the same wrist location and a very small difference in hand-joint location, yet Figure~\ref{fig:skeleton_restpose_wrong} shows a biologically unfeasible body pose in the mesh. Therefore, it is important to predict the 3D full-body mesh to represent the pose in addition to skeleton-based joint locations.


\subsection{3D Human Pose Representation}
In this paper, we represent the 3D human pose using the Skinned Multi-Person Linear (SMPL) model \cite{loper2015smpl}. We selected the SMPL representation because
\begin{enumerate*}
    \item it can represent varying human-body shapes and poses accurately and realistically \cite{loper2015smpl};
    \item the output is a full-body 3D mesh in addition to traditional skeleton-based 3D joint locations \cite{loper2015smpl, kanazawa2018end}; and
    \item it is a parametric statistical model that can easily represent the location, pose, and shape of a person by a vector of parameters.
\end{enumerate*}
The SMPL model has been used widely in image-to-pose estimation \cite{tan2017indirect, pavlakos2018learning, lassner2017unite}, yet few previous work exists on predicting/forecasting SMPL models into the future, particularly in global coordinate frames.

The SMPL model is formulated by three types of parameters, translation $\vec{\gamma}$, pose $\vec{\theta}$, and shape $\vec{\beta}$. The 3D body mesh is notated as
$M \left(\vec{\gamma}, \vec{\theta}, \vec{\beta}\right)$. The translation (``trans'') has three parameter values, indicating the global translation (distance in meters from the data capture system to the person) in x, y, and z axes. The pose parameters consist of the axis-angle representation of the relative rotation of 23 joints in a skeleton rig of the body and three root orientation parameters in x, y, and z axes (a total of 72 parameters) \cite{loper2015smpl}. The shape has 10 parameter values and indicates the body shape of the person. Under this formulation, the task of predicting a 3D human pose becomes that of predicting 85 (=3 +72 +10) SMPL parameters. 

\subsection{Gait Biomechanics}
\label{sec:gaitbio}
In addition to maintaining a feasible body pose (i.e., avoiding twists such as in Figure~\ref{fig:hand}), it is important to take the biomechanical characteristics of human gait into consideration. Gait analysis is a long-standing field of study and has had enormous impact on human locomotion and the development of bipedal robots \cite{vaughan1999dynamics, winter1995human, shajina2012human, felis2016synthesis}. For the specific task of pedestrian-walking pose prediction, we review related works in human gait studies and draw inspiration from three prominent biomechanical characteristics: mirror symmetry of human body, gait periodicity, and the change of ground reaction force in a human gait cycle in our network.

\begin{figure}[h!]
\centering
\begin{subfigure}{0.15\textwidth}
\includegraphics[scale=0.14, trim={5mm 5mm 0 0},clip]{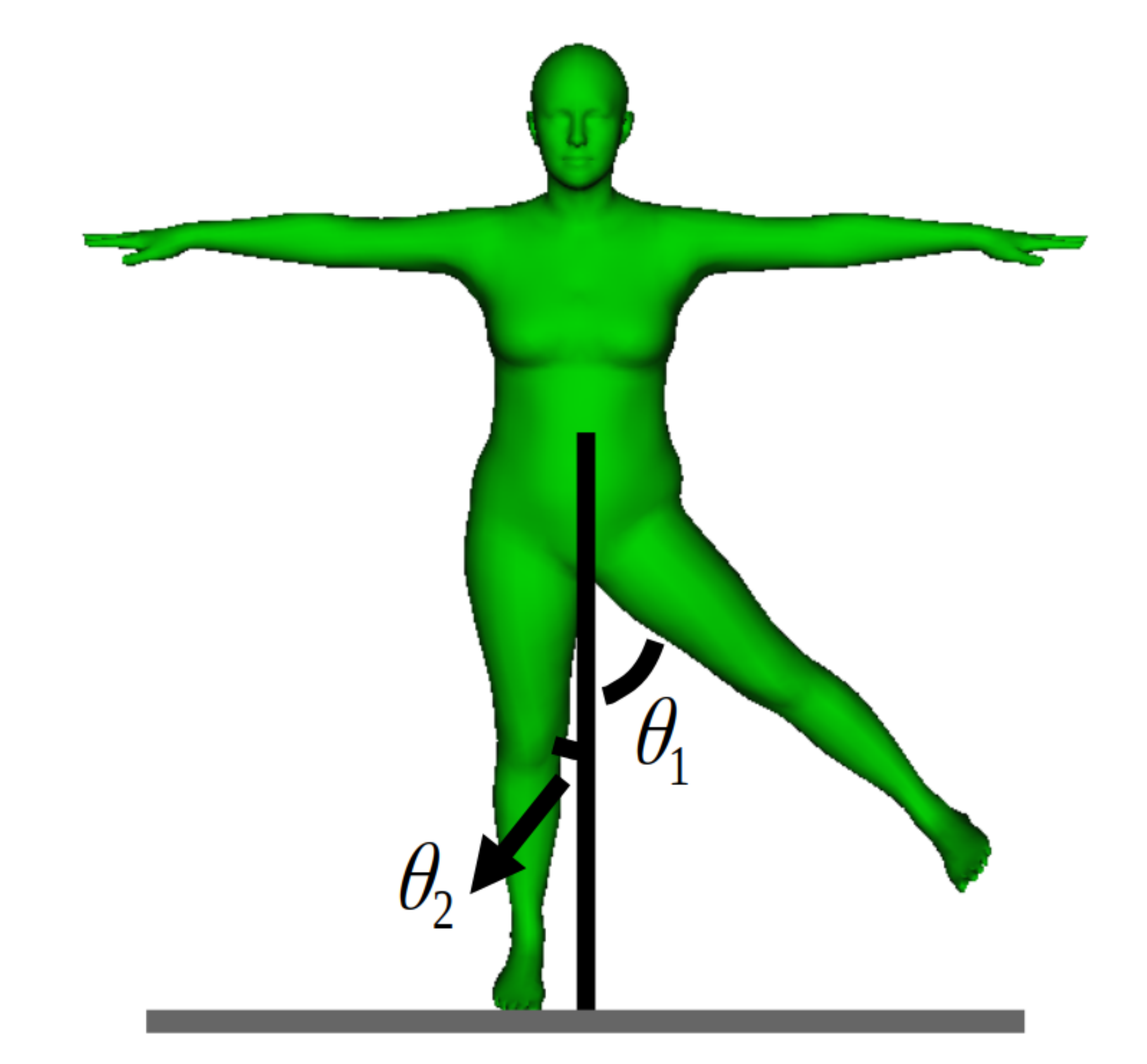}
\caption{ }
\label{fig:sym1}
\end{subfigure}
\begin{subfigure}{0.15\textwidth}
\includegraphics[scale=0.14, trim={0mm 0 0mm 0},clip]{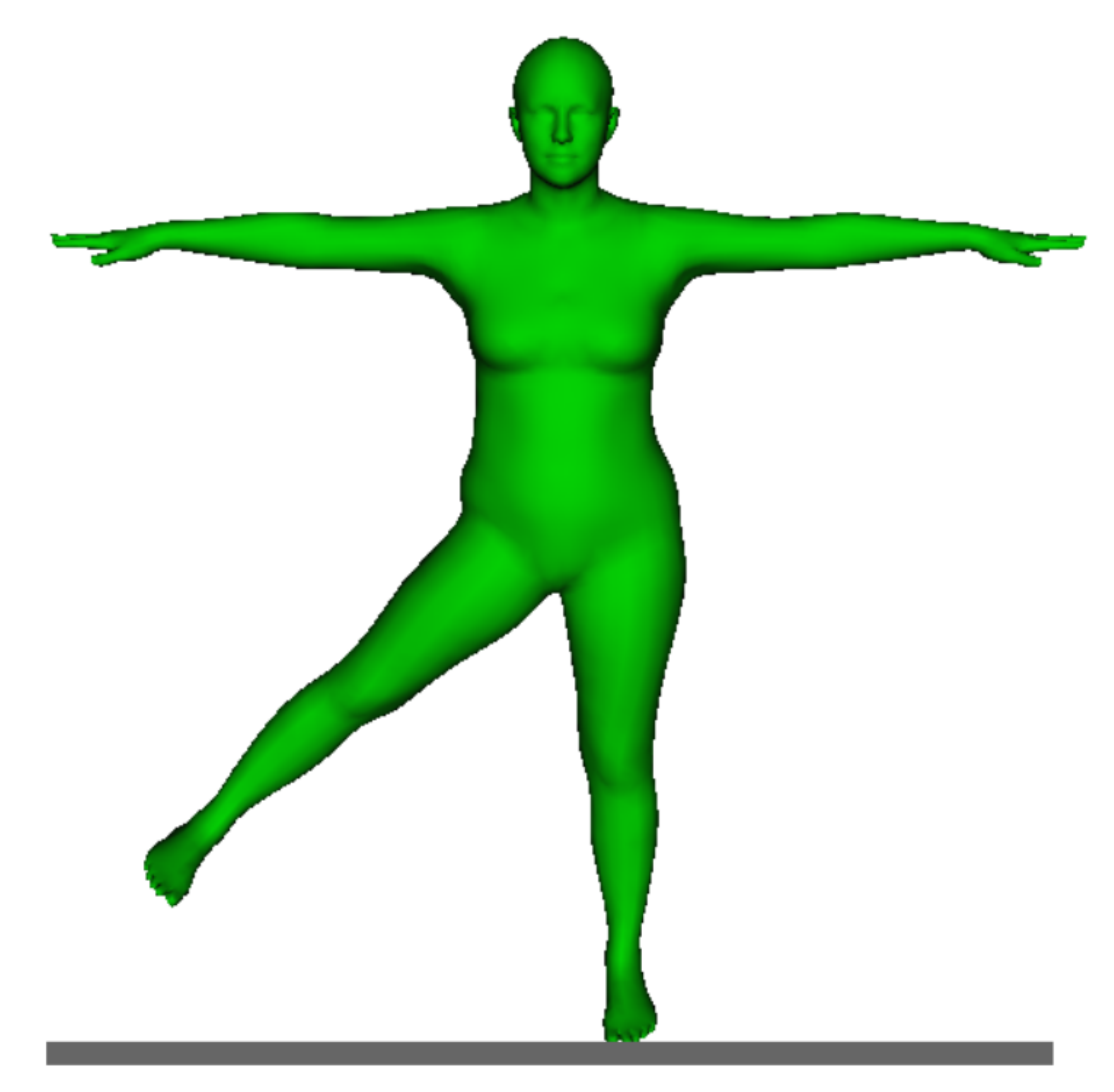}
\caption{ }
\label{fig:sym2}
\end{subfigure}
\begin{subfigure}{0.15\textwidth}
\includegraphics[scale=0.14, trim={0mm 0 0mm 0},clip]{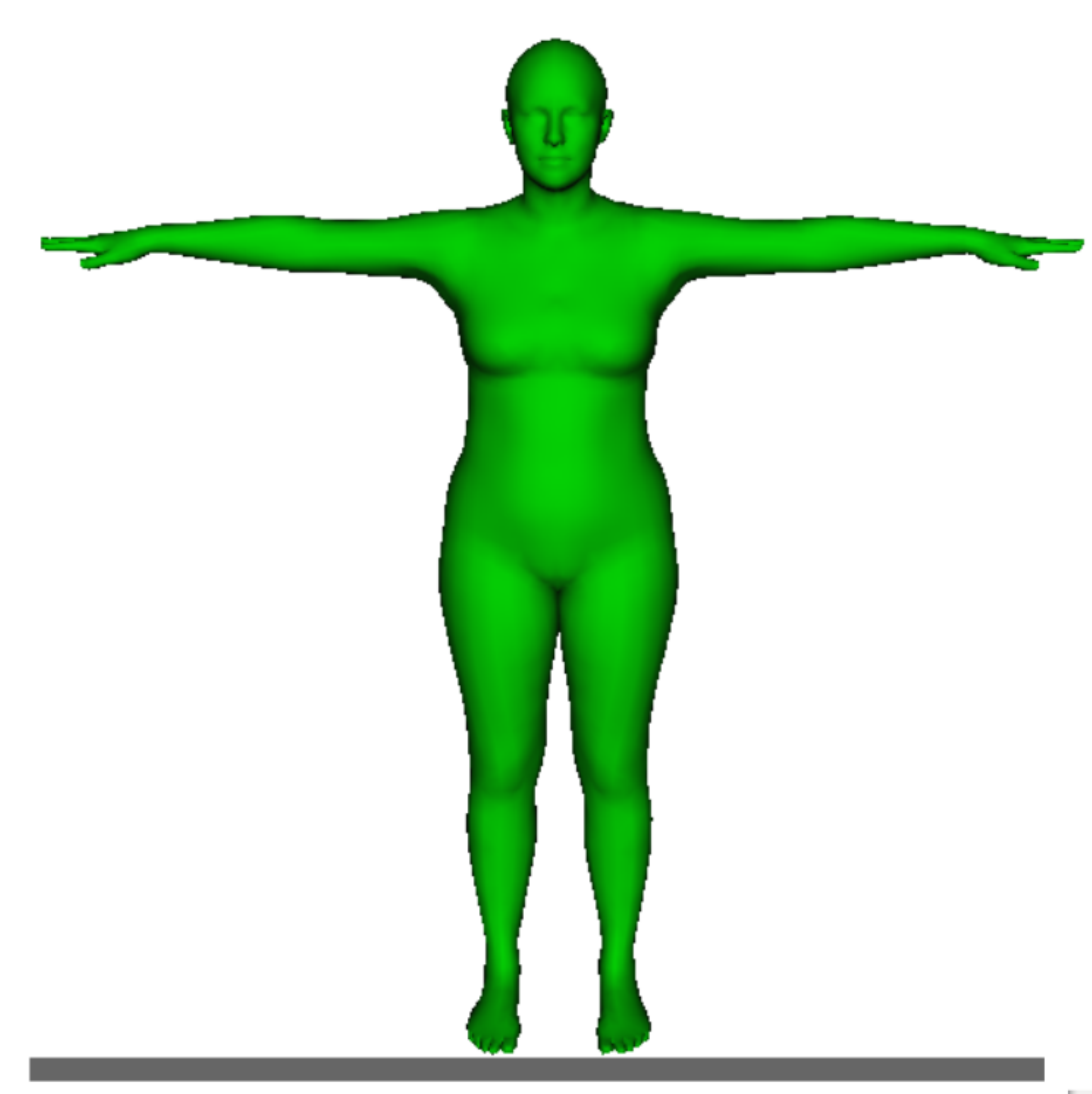}
\caption{ }
\label{fig:sym3}
\end{subfigure}
\caption{An illustration for human symmetry. $\theta_1$ is the angle between the left leg and the orthogonal line to the ground plane that runs through the center of the hip, and $\theta_2$ is the angle between the right leg and the center line. (a) An example when $|\theta_1| >|\theta_2| $. (b) An example when $|\theta_1| < |\theta_2| $. (c) An example when $|\theta_1| = |\theta_2| $. Among these three poses, pose (c) is the most stable and most similar to the natural human leg pose during standing/walking.}
\label{fig:symmetry}
\end{figure}

The bilateral/mirror symmetry of a healthy human body has long been observed in the literature \cite{mert2014walking, xu2009analysis, ankarali2015variability}. When the legs are positioned symmetrically along the center of the hip, the person is in balance. As shown in Figure~\ref{fig:symmetry}, it is desirable that $\theta_1 = -\theta_2 $   (also see rest pose in Figure~\ref{fig:skeleton_restpose_true}).  Similar symmetry can be observed for the two shoulder joints as well \cite{ramakrishnan2018comparing}. 

Cyclic leg movement is another important feature in human gait \cite{vaughan1999dynamics, yoo2003markerless}. It has been observed that humans walk with rhythmic and periodic motion \cite{sun2014human}. Step after step, a person's leg movement follows the cyclic motion with the assumption that all successive cycles are approximately the same as the first when traveling at a constant speed~\cite{vaughan1999dynamics}. In addition, it is assumed that the speed, stride, and direction during a normal walking cycle, and all successive cycles, do not suddenly change without an external force (e.g., a person does not suddenly flip \ang{180} during normal walking) \cite{mak2008sudden}. We observe such periodicity in our proposed network.

In addition, sufficient ground reaction forces (GRFs) are needed to support the body during walking \cite{vaughan1999dynamics}. The GRFs are applied through the feet, which means at least a part of one foot must be in contact with the ground \cite{vaughan1999dynamics}. To this end, we compute a local ground plane at the scene and map our body mesh prediction to ensure physically plausible contact between the feet and the ground.


\section{Method}
\label{sec:method}
\begin{figure}[h]
\centering
\includegraphics[width=0.5\textwidth, trim={0mm 0mm 0mm 2mm},clip]{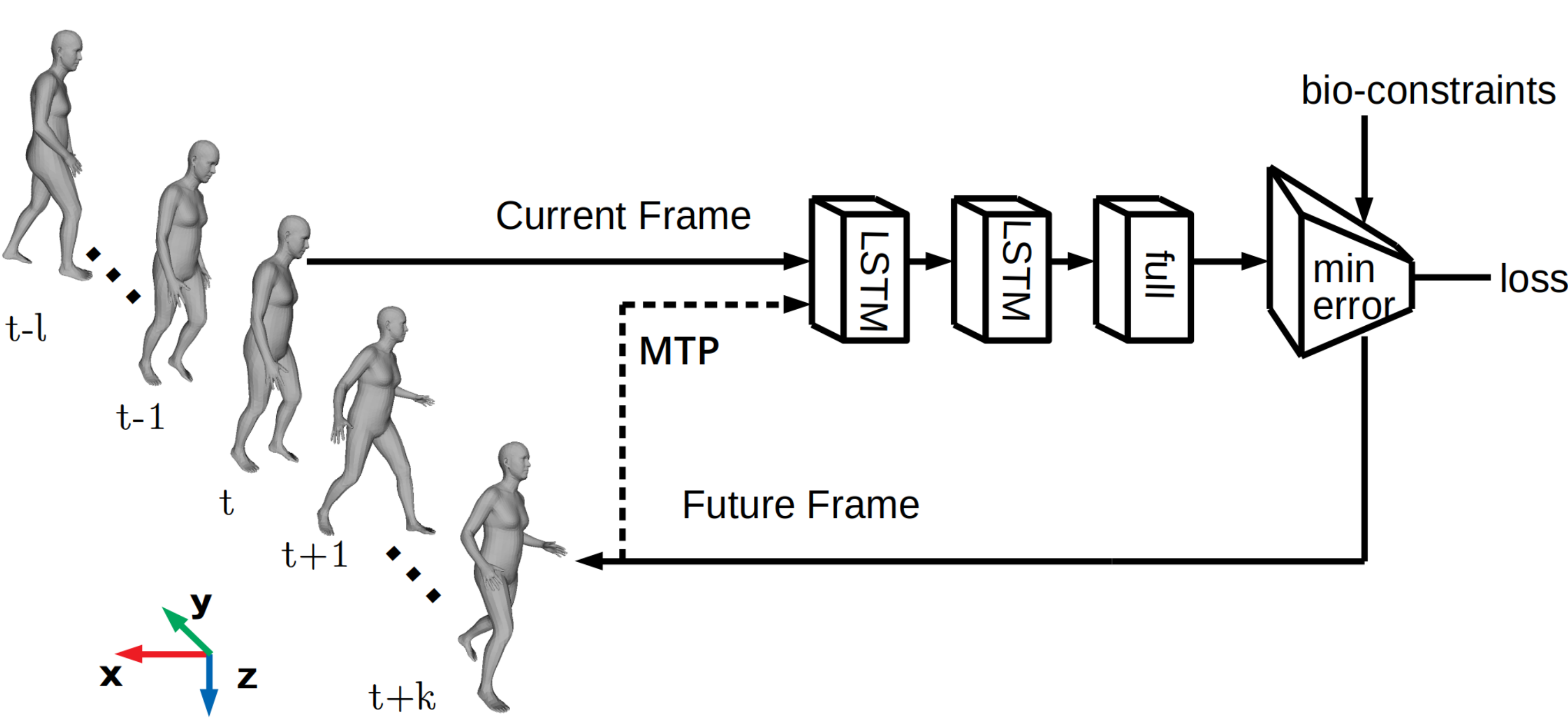}
\caption{An illustration for our proposed network. This illustration is inspired by the network diagram in \cite{vondrick2016anticipating} with the network architecture modified for our specific design. \textcolor{black}{The inputs and outputs of the network are vectors of SMPL parameters for all pedestrians in the scene. The bio-constraints were enforced through the training objectives in the network. For MTP, the predictions were continuously fed back to the network to predict all future timesteps.} }
\label{fig:lstm_pic}
\end{figure}

The goal of our network is to predict 3D full-body meshes in future frames, given 3D poses in past frames. Figure~\ref{fig:lstm_pic} illustrates the network diagram of our proposed approach. Details about the network architecture and error functions are described in the following subsections.

%


\subsection{Network Architecture}
\label{sec:networkarch}

We implemented a two-layer stacked LSTM recurrent neural network followed by a densely-connected neural network (NN) layer as our basic network architecture. This architecture was inspired by the LSTM-3LR method \cite{fragkiadaki2015recurrent}. We experimented with the number of layers (ranging from one to five) and found that the root mean square prediction error (RMSE) stopped decreasing after adding layer three in our experiments; therefore, we settled on a two-layer stacking architecture. We used this LSTM structure to predict both SMPL translation and pose parameters (3 translation parameters and 72 pose parameters, respectively). \textcolor{black}{We define $l$ as the look-back window length in the training sequences, $N$ is the total number of training sequences, and $q$ is the parameter dimensions ($q=3$ for translation and $q=72$ for pose parameters). Thus, the input size of the network is $(N, l-1, q)$. The $l-1$ dimension is because we use frame difference as part of our training objective functions, which will be further described in Section~\ref{sec:trainingobj}. } We assume the shape parameters (10 beta parameters) of each person remains the same as the previous frame (the person's body shape does not change from frame to frame). Each LSTM layer consists of 32 units (determined through experimentation). Section~\ref{sec:trainingobj} describes, in detail, our bio-inspired training objective function (the error module in Figure~\ref{fig:lstm_pic}). Section~\ref{sec:nfp} describes our procedure for next-frame prediction. Section~\ref{sec:mtp} describes our procedure for the multiple-timestep prediction (MTP).


\subsection{Training Objectives}
\label{sec:trainingobj}
We incorporate the three prominent biomechanical characteristics: gait periodicity, mirror symmetry of human body, and change of ground reaction force (GRF) in a human gait cycle in the training objectives of our network.


\begin{figure}[h]
\centering   
\includegraphics[width=0.5\textwidth, trim={0mm 0mm 0mm 2mm},clip]{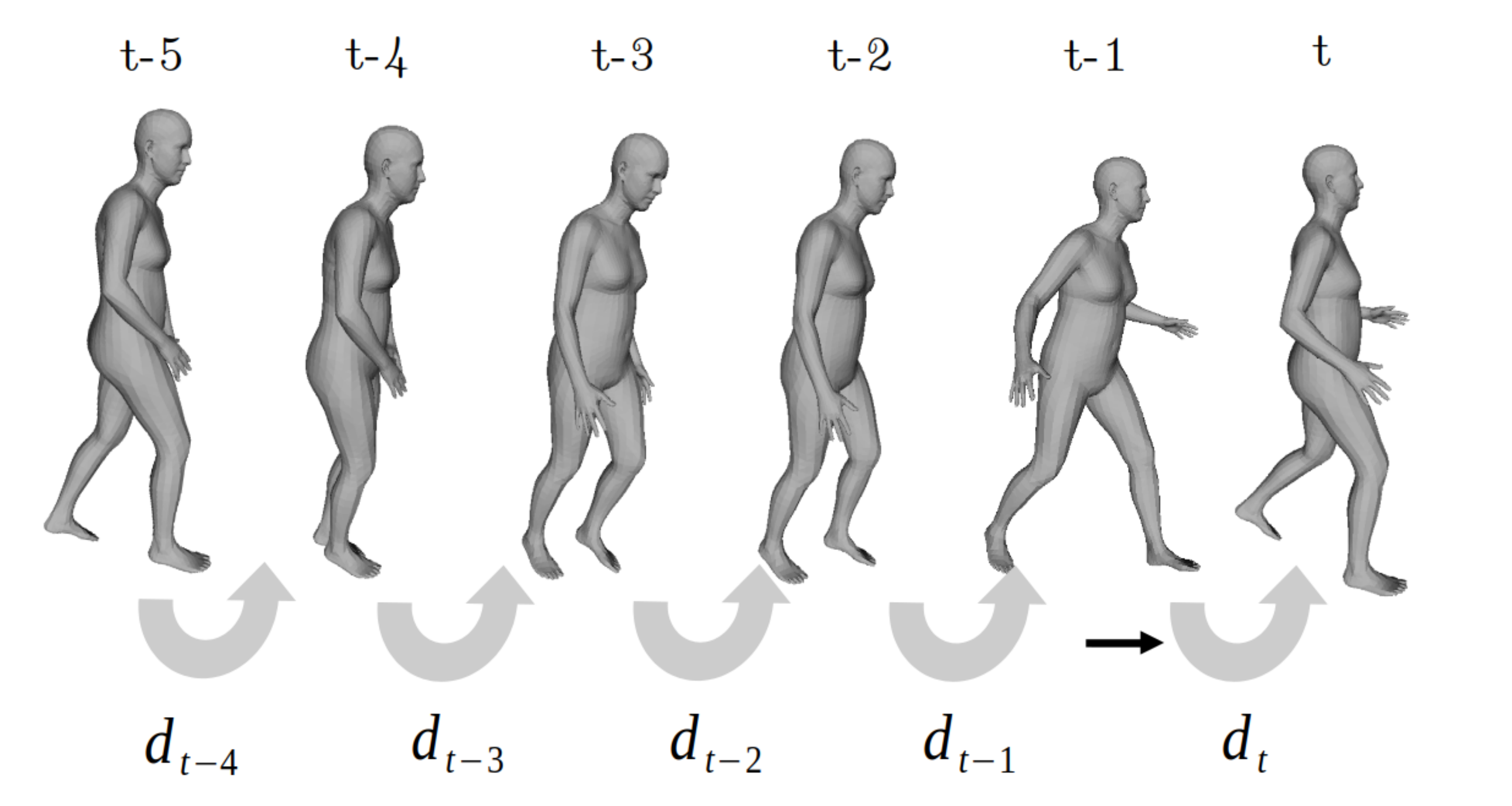}
\caption{An illustration for periodicity loss by predicting the frame difference (next-frame prediction).}
\label{fig:diff_pic}
\end{figure}

First, to address gait periodicity, we express the periodicity loss as the mean absolute error between the frame difference in the prediction sequence and the ``true'' frame difference in the training data. We illustrate the process (when $l=5$) in Figure~\ref{fig:diff_pic}. Given the translation and pose parameters for the last $l$ timesteps as $x_{t-5}, ..., x_{t-1}$, our goal is to predict translation and pose parameters for the next timestep $x_t$. Based on the assumption that the speed, stride, and direction do not suddenly change during walking cycles \cite{mak2008sudden}, we assume that the differences between frames remain steady. Also, the legs retain a cyclic motion. Therefore, we transform the problem into predicting the difference between frames. We define $d_t = x_t - x_{t-1}$ for the difference at timestep $t$. We then use $d_{t-4}, ..., d_{t-1}$ as inputs to our network and predict $\hat{d}_t$ as output. Then, our output translation and pose at time $t$ is given by $x_{t-1} + \hat{d}_t$. Thus, the periodicity loss $L_c$ for the sequence can be expressed as:
\begin{equation}
L_c = \left | d_t - \hat{d}_t \right |
\label{eq:Lc}
\end{equation}
%

Second, as discussed in Section~\ref{sec:gaitbio}, a person is stable when the left and right legs and shoulder joints are in mirror symmetry. Thus, we can write the loss based on body mirror symmetry as:
\begin{equation}
L_s = \left|\theta_{leg1} + \theta_{leg2} \right| + \left|\theta_{sho1} + \theta_{sho2} \right|,
\label{eq:Ls}
\end{equation}
where $\theta_{leg1} $ and $\theta_{leg2} $ are the angles between the left and right legs and the center vertical line at the upper thigh joints, and $\theta_{sho1} $ and $\theta_{sho2} $ are the angles between the left and right arms and the center vertical line at the shoulder joints.

\begin{figure}[h]
\centering   
\begin{subfigure}{0.12\textwidth}
\includegraphics[scale=0.2, trim={0 10mm 0mm 0},clip]{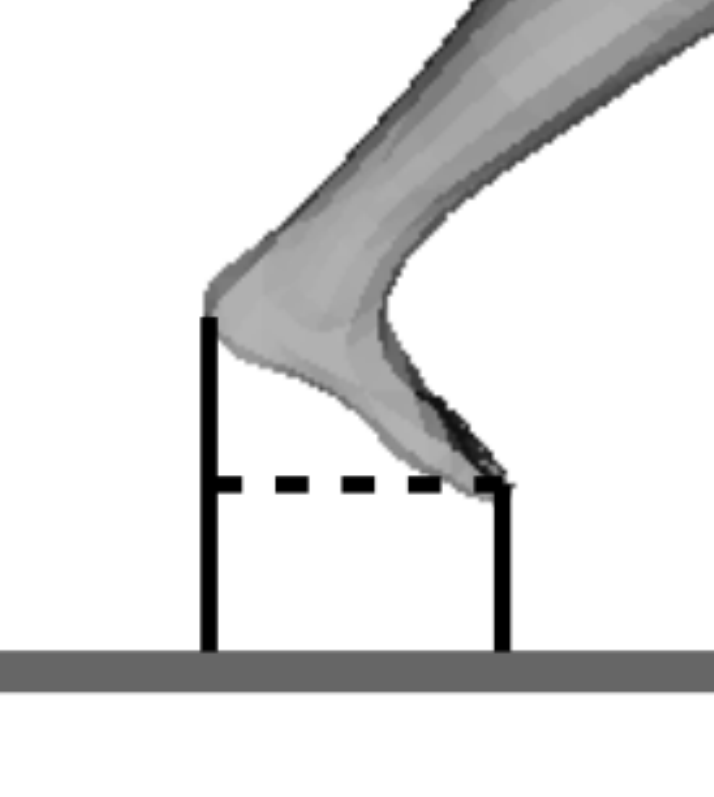}
\caption{ }
\label{fig:feet1}
\end{subfigure}
\begin{subfigure}{0.12\textwidth}
\includegraphics[scale=0.2, trim={0mm 10mm 0mm 0},clip]{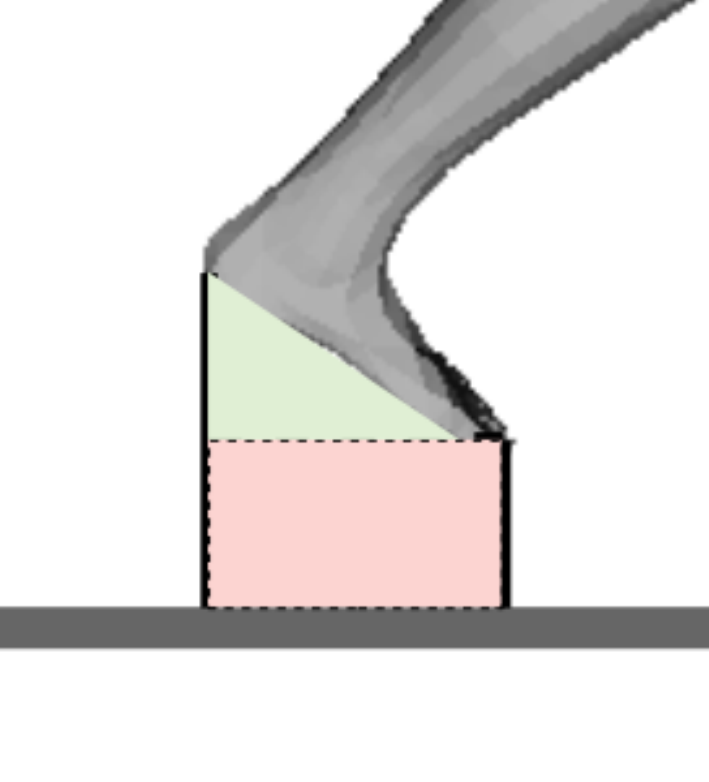}
\caption{ }
\label{fig:feet2}
\end{subfigure}
\begin{subfigure}{0.18\textwidth}
\includegraphics[scale=0.35, trim={0 12mm 0mm 0},clip]{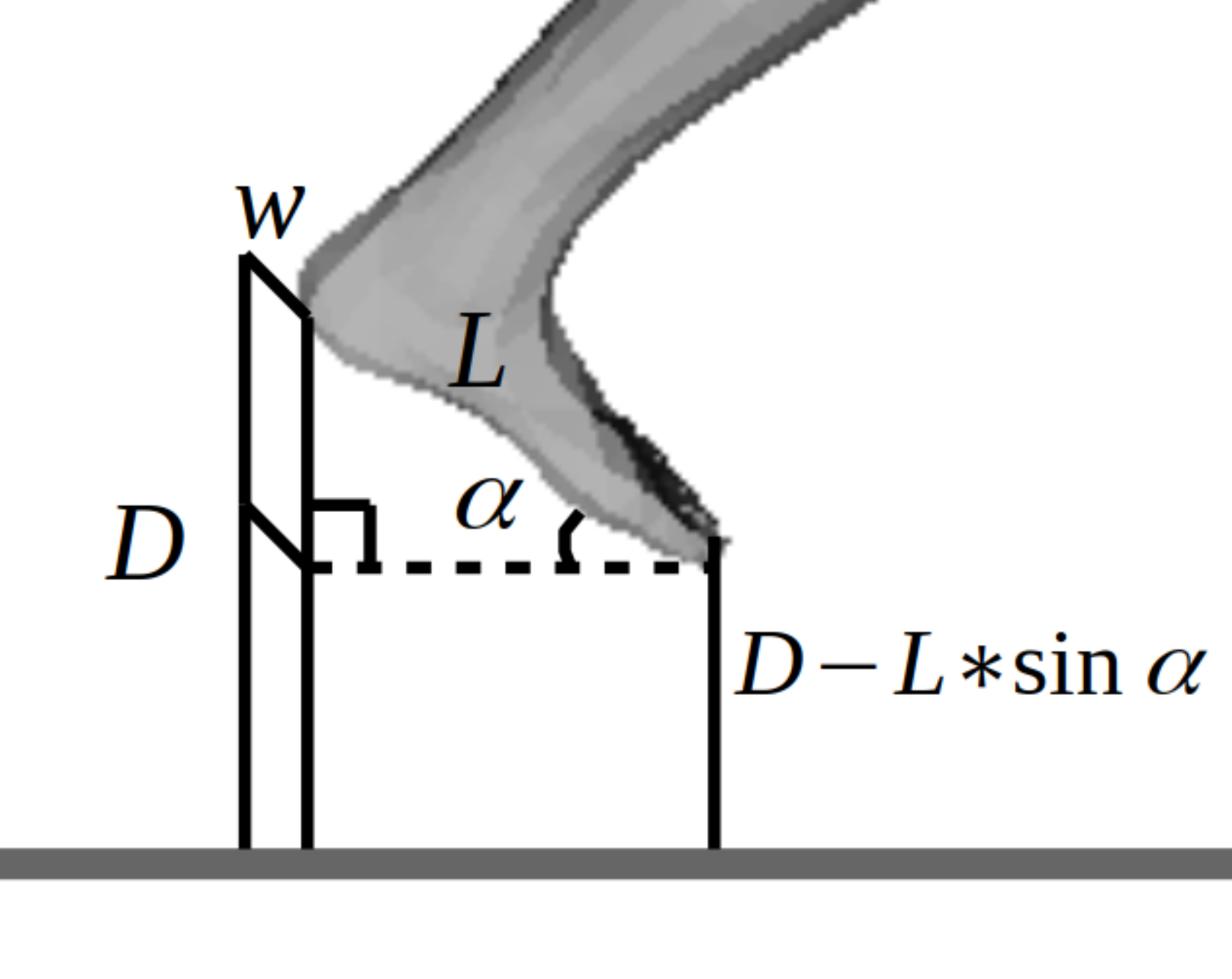}
\caption{ }
\label{fig:feet1}
\end{subfigure}
\caption{An illustration for the ground constraint on the feet. (a) 2D view of the space between the feet and ground. (b) Pink shade: rectangular cube; Green shade: a triangular prism. (c) With mathematical notations.}
\label{fig:feet_pic}
\end{figure}

Lastly, in order to provide sufficient ground reaction forces, we constrain the feet to the ground. Given ground elevation $G$ at each person's location in each frame, we minimize the volume between the feet and the ground, as shown in Figure~\ref{fig:feet_pic}. We simplify the volume model between the feet and the ground as the sum of the volumes of a rectangular cube (shaded in pink) and a triangular prism (shaded in green). We do so for both feet so, in sum, at least some transfer of force is occurring between the feet and the ground. We also encourage more ground contact-- humans generally use their full plantar aspect (the underside/sole of their feet) during walking and do not usually tiptoe \cite{phillips2017gait}. Thus, the volume loss from the ground plane is written as:
\begin{equation}
\resizebox{.44 \textwidth}{!} {
$\begin{split}
 L_g &= w*(D-L*sin\alpha)*(L*cos\alpha) + \frac{1}{2}  w * (L*sin\alpha)* (L*cos\alpha) \\
&= w*D*(L*cos\alpha) -  \frac{1}{2}  w * (L*sin\alpha)* (L*cos\alpha)
\end{split}$
}
\label{eq:Lg}
\end{equation}
where $w$ is the width of the human foot, $D$ is the vertical distance between the heel of the foot to the ground, $L$ is the length of the human foot, and $\alpha$ is the angle between the foot and the horizontal ground plane. Note this requires a local ground plane estimate. In our case, this is derived from LiDAR data from an Autonomous Vehicle (AV), but could also be estimated from stereo or other monocular vanishing point cues~\cite{zhao2007global, micusk2008towards, gardner2010vertical}. The $w$, $D$, and $L$ values are estimated from the SMPL rest pose.


Therefore, our training objective (total loss function) can be written as:
\begin{equation}
\min L = L_{c} + \lambda_1 L_{s} + \lambda_2 L_{g}
\label{eq:loss}
\end{equation}
where $L_c$ is the loss from the gait cycle, $L_s$ is the loss based on body mirror symmetry, $L_g$ is the loss based on volume from the ground plane, and $\lambda_1$ and $\lambda_2$ are user-set regularization parameters to adjust the weighting of bio-inspired loss function terms. In our following experiments, we set $\lambda_1=10$ and $\lambda_2=0.01$ (determined through loop testing).


\subsection{Next-Frame Prediction}
\label{sec:nfp}
We formulate the next-frame prediction as a supervised learning problem. First, we construct training, validation, and testing sequences by creating batches from all pose sequences of length $l+1$, which we denote by $\left \{ \vec{X}_{t-l}, ..., \vec{X}_{t-1}, \vec{X}_t \right \}$ for all $t$ in the dataset. The first $l$ poses were the inputs to the network and the last is the next-frame target to be predicted. When $l>1$, we use the the proposed 2LR-LSTM network with training objective \eqref{eq:loss} for prediction. When $l=1$ (only given one frame to predict the next), we define the frame difference to be the median frame difference in all training data and apply such frame difference to predict the next frame, assuming that a person follows the leading direction of the population flow \cite{zhao2017unified}.


\subsection{Multiple Timestep Prediction}
\label{sec:mtp}
In multiple-timestep prediction, given $\left \{ \vec{X}_{t-l}, ..., \vec{X}_{t-1} \right \}$, we first predict $\vec{X}_t$. Then, this prediction at time $t$ is fed back to the network and we predict the pose at $t+1$ based on the sequence  $\left \{ \vec{X}_{t-l+1}, ..., \vec{X}_{t} \right \}$. This process is marked as ``MTP'' (dashed line) in Figure~\ref{fig:lstm_pic}. In this way, we can continuously output poses at time $t$, $t+1$, $t+2$, ..., $t+k$ for any timestep $k$ in the future.

\section{The PedX Dataset and Experimental Setup}
\label{sec:setup}
This section first describes the PedX dataset, the in-the-wild pedestrian pose dataset used for the experiments. Then, the baseline methods used for comparison and the evaluation metrics are described. The data pre-processing procedure for the PedX dataset is also presented.


\subsection{The PedX Dataset}
The PedX dataset \cite{kim2018pedx} was collected in 2017 in real urban intersections in downtown Ann Arbor, Michigan, USA. The dataset contains collections from three four-way-stop intersections with heavy pedestrian traffic. The PedX dataset contains over 10,000 pedestrian poses and over 1800 continuous sequences of varying length (average sequence length is six frames). The PedX dataset consists of data from two stereo RGB camera pairs and four Velodyne LiDAR sensors. The camera videos were collected at approximately six frames per second (fps). We collected this dataset from a parked car facing the intersection and recorded in-the-wild pedestrian behavior (pedestrians span a range of 5-45 meters from the cameras). The 3D pedestrian pose in each frame was obtained by optimizing the manually-annotated 2D pedestrian pose and 3D LiDAR point cloud as described in Kim et al.~\cite{kim2018pedx}. Given such (known) 3D pedestrian poses (also called ``3D training labels'') in a few frames in past sequences, our proposed network predicts the 3D pedestrian pose in the next frame and multiple timesteps in the future. 


The PedX dataset also contains an evaluation dataset collected and annotated in a controlled outdoor environment with a motion capture (mocap) system (named ``mocap dataset''). The mocap dataset was collected using the same setup as with the intersection data, but only contains one pedestrian with mocap markers. We evaluate the performance of our proposed method on the mocap dataset also, since the mocap ground-truths were available \cite{kim2018pedx}.


\subsection{Baseline Methods}
We compare our proposed bio-LSTM network with several baseline methods. We first compare our network with the two-layer stacked LSTM recurrent neural network followed by a densely-connected NN layer as a state-of-the-art baseline method (denoted the ``2LR-LSTM'' method as described in \cite{fragkiadaki2015recurrent}) without the bio-constraints. The standard 2LR-LSTM is trained on 1) the skeleton-based 3D joint locations (denoted ``skeleton joints'' in the following tables) and 2) the SMPL parameters directly (denoted ``trans.+pose''). The input size of this baseline network is $(N, l, q)$, as defined in Section~\ref{sec:networkarch}.  

Then, we compare our work with the ``frame difference'' baseline method \cite{zhan2007improved}. In this baseline method, 3D pedestrian poses are predicted by computing the difference in translation and pose parameters in the past frames and then applying this difference to future frames. For example, as shown in Figure~\ref{fig:diff_pic}, we compute $d_0 = median \left\{ d_1, d_2, ... d_{l-1}\right\}$. Then, the predicted translation and pose at $t$ equals translation and pose at $t-1$ plus $d_0$. This baseline method essentially enforces the $L_c$ constriant, but does not train an LSTM network. 

In addition, we analyze the effect of each loss term in our bio-inspired objective function and summarize the results for using different loss terms in an ablation study.


\subsection{Evaluation Metrics}
The outputs of our proposed bio-LSTM network are 85 SMPL parameters. \textcolor{black}{Note that we assume the 10 shape parameters do not change from frame to frame for each person.} From the SMPL parameters, we compute the locations of the 6890 vertices that forms the 3D full-body mesh, according to Loper et al. \cite{loper2015smpl}. In this paper, we evaluate our method using the vertex root-mean-square error (vertex RMSE) as well as the standard 3D mean-per-joint-position error (MPJPE) \cite{ionescu2014human3, kim2018pedx}. As the MPJPE only evaluates skeleton-based joint locations and does not capture  differences in the full mesh, vertex RMSE is helpful in evaluating biologically infeasible poses such as Figure~\ref{fig:hand}.  We also computed the RMSE in global translation as well as the mean-per-joint-angular error (MPJAE) \cite{von2018recovering} on all 24 joint angles.


\subsection{Data Pre-Processing}
\label{sec:preproc}
In our prediction experiments, we normalize the translation and pose parameters. The translation parameters are normalized by their max and min ranges in x, y, and z axes, and the joint angle magnitudes are normalized between $[0, 2\pi)$. In our PedX experiments, we use 85\% of data sequences as training, 10\% of data sequences as validation, and 5\% of data sequences as testing. This split scheme was selected to ensure a large number of training sequences as well as enough test data to evaluate our results. The sequences were randomly shuffled during training and we report the mean and standard deviation across three random initializations. 

\begin{figure}[h!]
\centering   
\begin{subfigure}{0.32\textwidth}
\includegraphics[width=\textwidth, trim={0 0mm 0mm 0},clip]{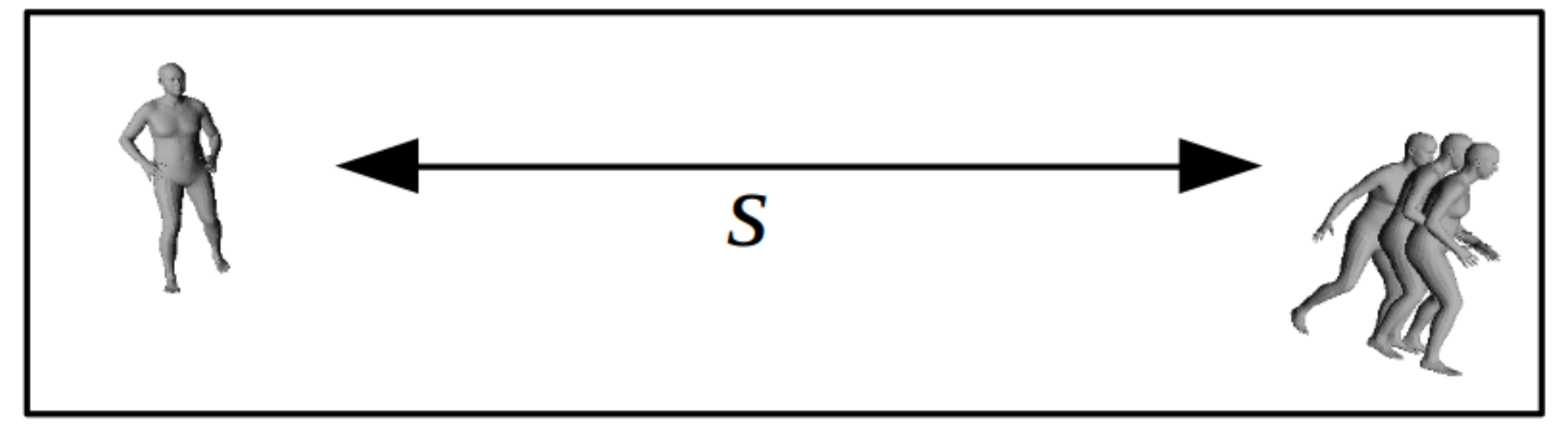}
\caption{ }
\label{fig:wrong_model_1}
\end{subfigure}
\begin{subfigure}{0.16\textwidth}
\includegraphics[width=\textwidth, trim={0mm 0mm 0mm 0},clip]{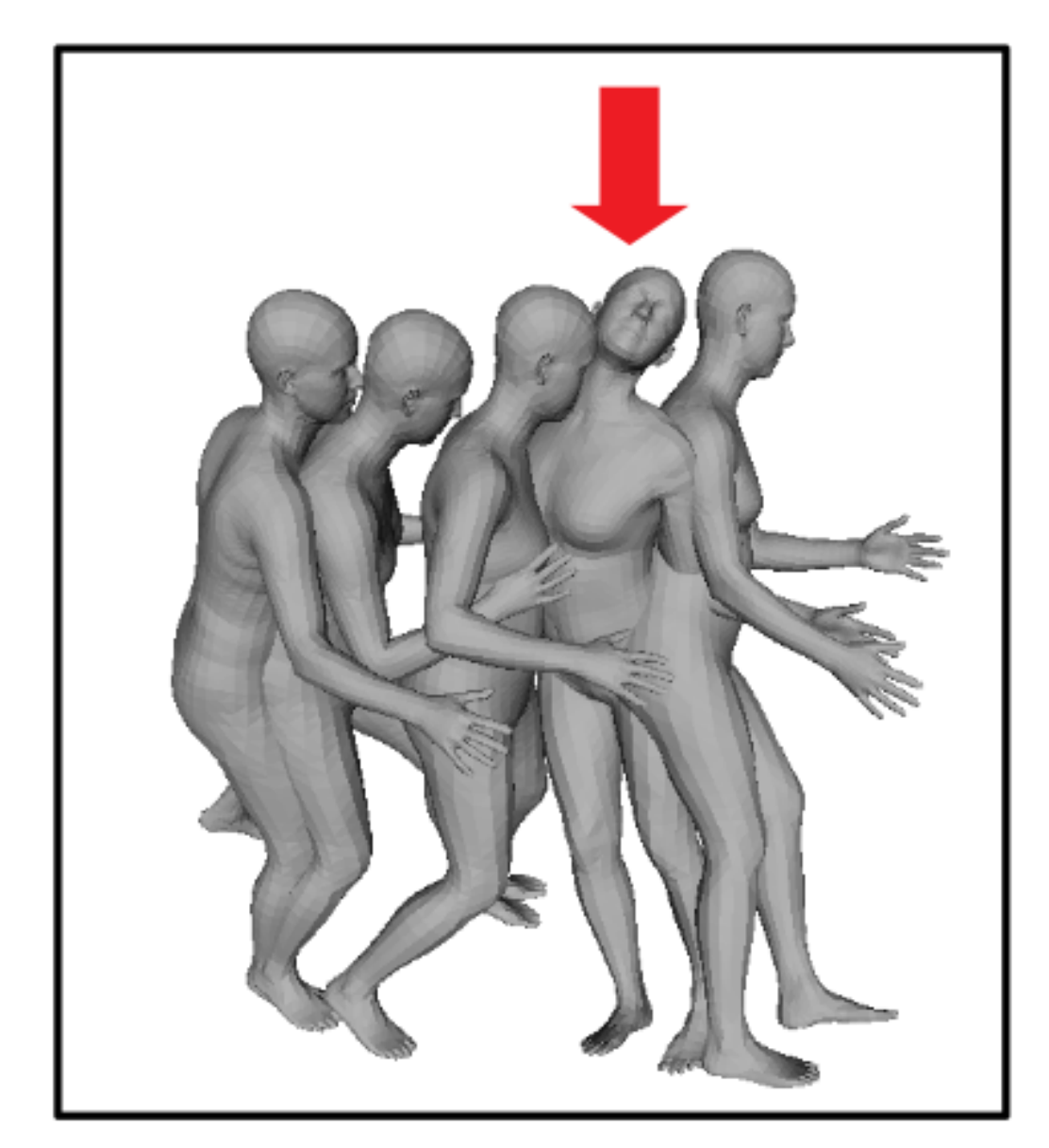}
\caption{ }
\label{fig:wrong_model_2}
\end{subfigure}
\caption{An illustration of noisy poses (``outliers'') from the field data. (a) A person model with a sudden jump in translation. Currently, the translation distance threshold $s=0.6m$. (b) A person model with wrong body orientation in a sequence (marked with red arrow).}
\label{fig:wrong_model}
\end{figure}

Our training labels came from the previous 3D pose estimation optimization method of Kim et al. \cite{kim2018pedx} Although their method achieves state-of-the art estimation results, there is still noise when capturing data from the vehicle 
due to measurement inaccuracy in the 3D LiDAR point cloud data and the long observation range. 
In our prediction experiments, we eliminated noisy models (``outliers''), such as frames with large distance within a sequence or sudden change of root orientation, as shown in Figure~\ref{fig:wrong_model}. We present  our prediction results trained on both filtered and noisy labels to show that our proposed method can handle such noise robustly.


\section{Results and Discussion}
\label{sec:results}
In this section, we present results for next-frame prediction on both the PedX and the mocap data. 
 The standard deviation across three random initializations are presented in parentheses in the following tables.

\begin{figure}[h]
\centering   
\includegraphics[width=0.5\textwidth, trim={0 10mm 0mm 0},clip]{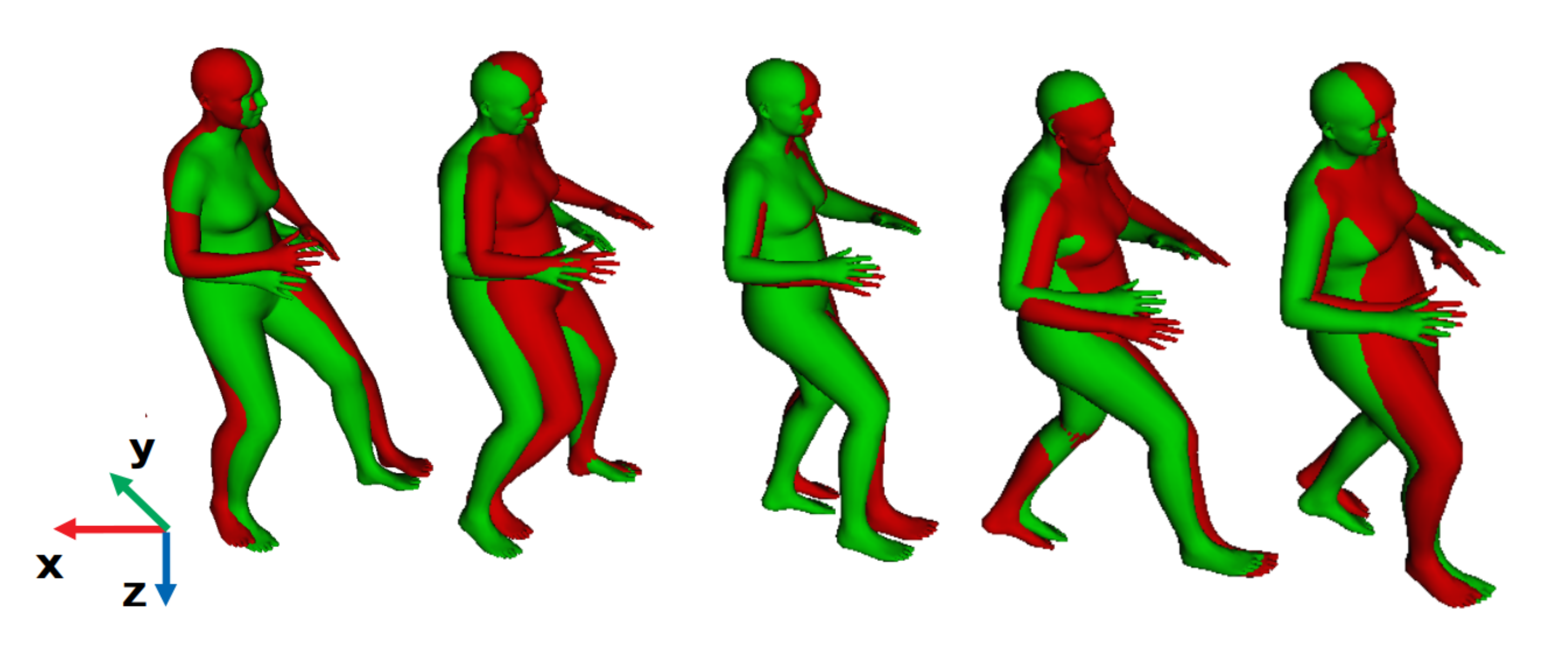}
\caption{A qualitative example for predicted pedestrian 3D poses in a walking cycle. The green meshes are predicted by our network and the red meshes are the ``ground truth'' labels optimized in \cite{kim2018pedx}.}
\label{fig:predresults_single_ex1}
\end{figure}

Table~\ref{table:pedxresults} presents results on next-frame prediction on the PedX dataset with look-back window length $l=5$. The $l$ value was chosen as a pedestrian generally completes a walking cycle in 5-6 frames in the PedX dataset. Table~\ref{table:mocapresults} presents results on next-frame prediction on the mocap dataset with $l=5$. Our method is able to achieve around 85mm error (full-body mesh in global frame) in outdoor intersection data and 73mm error in mocap data, with the global translation range of approximately 45 meters (thus, an error rate of 0.16\%-0.19\%). The average angle error is 13.5$^\circ$. Furthermore, in both experiments, our proposed network yields better prediction results (lower RMSE error) in translation, joints, vertex, and angle. We observed that the gait periodicity loss ($L_c$) was the most prominent feature and produced much smaller error compared with baseline methods (36.8\% improvement over predicting only skeleton joints and 21.0\% improvement on vertex RMSE).  Adding the mirror symmetry constraint ($L_s$) enabled modest performance gain (around 1.6\%). Figure~\ref{fig:predresults_single_ex1} shows a qualitative example of our prediction results.


\begin{table}[h!]
  \centering
\caption{Next-Frame prediction results on the PedX dataset, $l=5$. In all tables, the unit for trans, MPJPE, and vertex error is $\times10^{-3}m$.}
\label{table:pedxresults}
 \resizebox{\columnwidth}{!}{
\begin{tabular}{lcccc}
    \hline
\textbf{Methods}&\textbf{trans}  &\textbf{MPJPE} &\textbf{vertex  } &\textbf{MPJAE($^\circ$) } \\
\hline 
Skeleton Joints & -- & 130.8(18.1)  & -- & --  \\ 
Trans.+Pose & 81.6(18.9) &  102.2(16.8)& 104.4(15.4) & 16.1(1.8)  \\
Frame Diff. & 61.6(3.2)  & 109.6(10.5) & 107.8(9.79) & 23.9(3.4)\\ 
\textbf{Ours}($L_c$) & \textbf{52.9(2.7)} & \textbf{82.6(5.7)} & 85.2(5.4) &  \textbf{15.8(2.0)}    \\ 
\textbf{Ours}($L_c+L_s$) & 53.0(2.8) & \textbf{82.6(5.6)} & \textbf{84.8(5.1)} & \textbf{15.8(1.8)}  \\ 
\hline
  \end{tabular}
}
\end{table}

\begin{table}[h!]
  \centering
  \caption{Next-Frame prediction results on the mocap dataset, $l=5$.}
\label{table:mocapresults}
  \resizebox{\columnwidth}{!}{
  \begin{tabular}{ lcccc}
    \hline
\textbf{Methods}&\textbf{trans}  &\textbf{MPJPE} &\textbf{vertex} &\textbf{MPJAE($^\circ$)} \\
\hline 
Skeleton Joints & -- & 182.9(42.0)  & -- & --  \\ 
Trans.+Pose & 73.8(27.6) & 101.1(23.2) & 108.1(21.6) & 15.5(0.2)  \\
Frame Diff. & 72.3(23.3) & 84.8(15.5) & 87.1(15.0) & 11.3(1.4) \\
\textbf{Ours}($L_c$) & 48.8(1.2) & 68.2(0.8) & 73.8(0.7) & 11.3(0.1) \\ 
\textbf{Ours}($L_c+L_s$) & \textbf{48.6(1.2)}  & \textbf{67.4(1.3)} & \textbf{72.6(1.3)} & \textbf{11.2(0.0)}  \\ 
\hline
  \end{tabular} 
}
 \end{table}


\begin{table}[h!]
  \centering
 \caption{Next-Frame prediction results on the PedX and mocap dataset, $l=1$.}
\label{table:pedxmocapresults}
  \resizebox{\columnwidth}{!}{ 
\begin{tabular}{ lcccc}
    \hline
\textbf{Methods}&\textbf{trans}  &\textbf{MPJPE} &\textbf{vertex} &\textbf{MPJAE($^\circ$)} \\
\hline
Trans.+Pose on PedX & 158.8(19.1) & 180.0(19.2) & 172.1(19.4) &  \textbf{16.2(1.3)} \\
\textbf{Ours} on PedX & \textbf{144.6(10.2)}  & \textbf{165.4(10.9)} & \textbf{164.8(10.3)} &  19.9(1.8) \\ 
Trans.+Pose on mocap & 107.7(18.4) & 130.4(18.4) & 132.5(15.4) & 17.0(0.5)  \\ 
\textbf{Ours} on mocap & \textbf{77.8(13.4)}  & \textbf{91.0(9.9)} & \textbf{93.2(9.7)}  & \textbf{11.5(1.0)}  \\ 
\hline
  \end{tabular}  
}
 \end{table}

\begin{table}[h!]
  \centering
  \caption{Next-Frame prediction results on the mocap data, trained with noisy labels from PedX dataset, $l=5$.}
\label{table:pedxresults_noisy}
  \resizebox{\columnwidth}{!}{
\begin{tabular}{ lcccc}
    \hline
\textbf{Methods}&\textbf{trans}  &\textbf{MPJPE} &\textbf{vertex} &\textbf{MPJAE($^\circ$)} \\
\hline 
Skeleton Joints & -- &  292.8(32.2) & -- & --  \\ 
Trans.+Pose & 223.7(26.9) & 231.7(26.6) & 236.0(25.4) &  17.0(0.5) \\ 
Frame Diff. & 87.6(5.7) & 95.6(5.4) & 97.7(5.3)  & \textbf{ 10.7(0.5)} \\ 
\textbf{Ours} ($L_c$) & 65.7(0.1) & 82.5(0.3) & 85.6(0.3) & 11.7(0.1)  \\
\textbf{Ours} ($L_c+L_s$) & \textbf{65.6(0.3)} & \textbf{ 80.1(0.6)} & \textbf{ 83.3(0.7)} & 10.8(0.0)  \\ 
\hline
  \end{tabular}
}
 \end{table}


Table~\ref{table:pedxmocapresults} shows the prediction results when $l=1$, i.e., prediction without pose information from prior frames. From the table we can observe that our method still outperforms the standard 2LR-LSTM prediction without biomechanical constraints. The error is significantly smaller in mocap data than in the PedX data, as there is only one pedestrian in the mocap data and the frame difference is more regular than the in-the-wild intersection data with multiple pedestrians.


Table~\ref{table:pedxresults_noisy} shows the prediction results on mocap data, using models trained with noisy training labels that reflect the errors one typically seeing in real field data, as described in Section~\ref{sec:preproc}. As can be seen, the baseline methods have significantly higher error due to noise in the input data, yet our proposed methods yield almost comparable prediction results.

Table~\ref{table:groundresults} shows the ground distance error of our prediction results on the PedX data. We compare ground distance of our prediction results before and after adding the $L_g$ loss term. We also report the ground distance error from the previously estimated poses \cite{kim2018pedx}. It can be seen that the $L_g$ loss term was able to constrain the feet closer to the local ground plane. The remaining error is likely due to an estimation error of the local ground plane from the LiDAR point cloud data, and the simplified volume loss model in \eqref{eq:Lg}. The estimated lengths and widths of the foot and leg also change slightly due to the human body shape, which may contribute to the ground distance error as well. 


\begin{table}[h!]
  \centering
  \caption{Distance to the ground after adding $L_g$.}
\label{table:groundresults}
  \resizebox{\columnwidth}{!}{
  \begin{tabular}{ lc}
    \hline
\textbf{Methods} &\textbf{distance to the ground($\times10^{-3}m$)}   \\
\hline
Existing model \cite{kim2018pedx} & 32.4(0.9) \\
Ours($L_c+L_s$) &   40.1(2.1) \\ 
Ours($L_c+L_s+L_g$)  & \textbf{29.5(4.0)} \\ 
\hline
  \end{tabular}
}
\end{table}

\textcolor{black}{Table~\ref{table:pedxresults_action} shows the vertex RMSE results across different actions evaluated on a subset of annotated PedX dataset. The actions under investigation include:  simply walking (1307 sequences), carrying a coffee cup in the right hand (10 sequences), carrying something in the left hand (53 sequences), carrying/playing with cellphones (58 sequences), pushing a bicycle (45 sequences), and cycling (9 sequences). The mean and standard deviation values (in parentheses) were reported across sequences. As can be seen, our method was able to achieve lower vertex RMSE in all actions except for cycling. Among all the actions, simply walking has the lowest vertex RMSE, which is as expected since our objective functions were focused on walking gait and there are a large number of walking sequences in the dataset. The large error in cycling action was partly due to the limited number of frames in the dataset and that cycling and walking have significantly different stride. In this case, our network (which was trained for pedestrian walking poses) was not able to predict as well for cyclists, while the frame difference method was able to preserve more accurate cycling stride (i.e., the translation difference). However, it is worth noting that, even with limited cycling training data, our network is still able to predict biomechanically feasible poses for cyclists.\footnote{Additional mesh prediction results and examples can be viewed in our supplementary video. \textsuperscript{\textcopyright} IEEE Xplore Digital Library Link is: \url{https://ieeexplore.ieee.org/document/8626436/media\#media}.\label{footnote}}}


\begin{table}[h!]
  \centering
\caption{\textcolor{black}{Vertex RMSE results ($\times 10^{-3}m$) on different actions in the annotated PedX dataset, $l=5$. }}
\label{table:pedxresults_action}
 \resizebox{\columnwidth}{!}{
\begin{tabular}{lccc}
    \hline
\textbf{Actions}&\textbf{simply walking}  &\textbf{cup} &\textbf{carry (left hand)} \\\hline
{Frame Diff.} & 97.8(37.6) & 108.3(16.9)& 92.6(30.0) \\
{Ours} & \textbf{73.2(30.9)} & \textbf{90.0(21.3)}& \textbf{73.7(32.6)} \\
\hline
\textbf{Actions} &\textbf{phone} &\textbf{push bike}  &\textbf{cycling} \\\hline
{Frame Diff.}  & 110.0(37.3) & 97.4(38.9) & \textbf{119.0(31.5)}\\
{Ours} & \textbf{91.2(33.3)} & \textbf{85.7(33.3)} & 331.1(66.5)\\
\hline
  \end{tabular} 
}
\end{table}

\begin{figure}[h!]
\centering
\begin{subfigure}{0.22\textwidth}
\includegraphics[width=\textwidth, trim={0 0mm 7mm 27mm},clip]{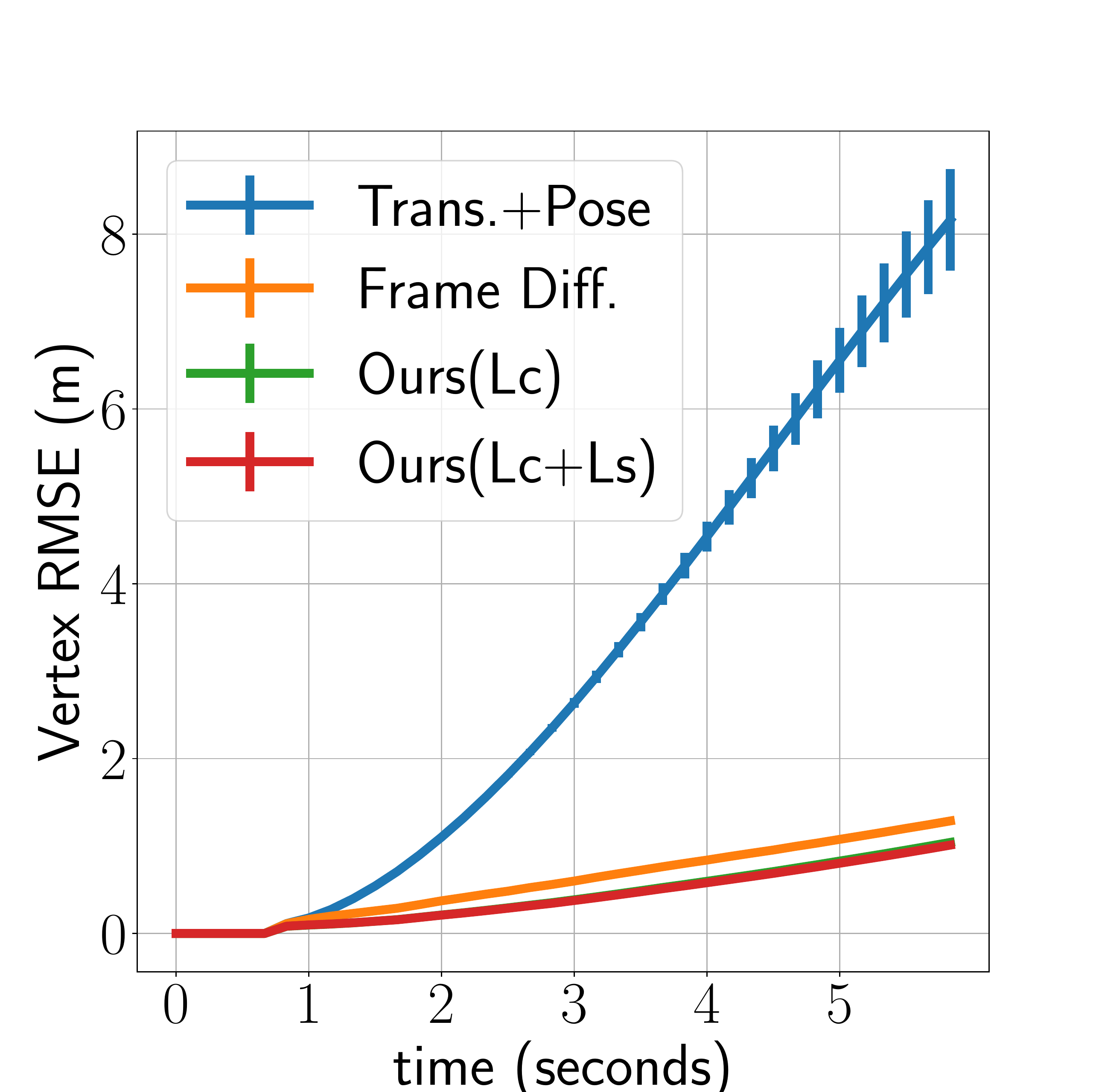}
\caption{ }
\label{fig:MTP_vertex}
\end{subfigure}
\begin{subfigure}{0.26\textwidth}
\includegraphics[width=\textwidth, trim={2mm 0mm 0mm 27mm},clip]{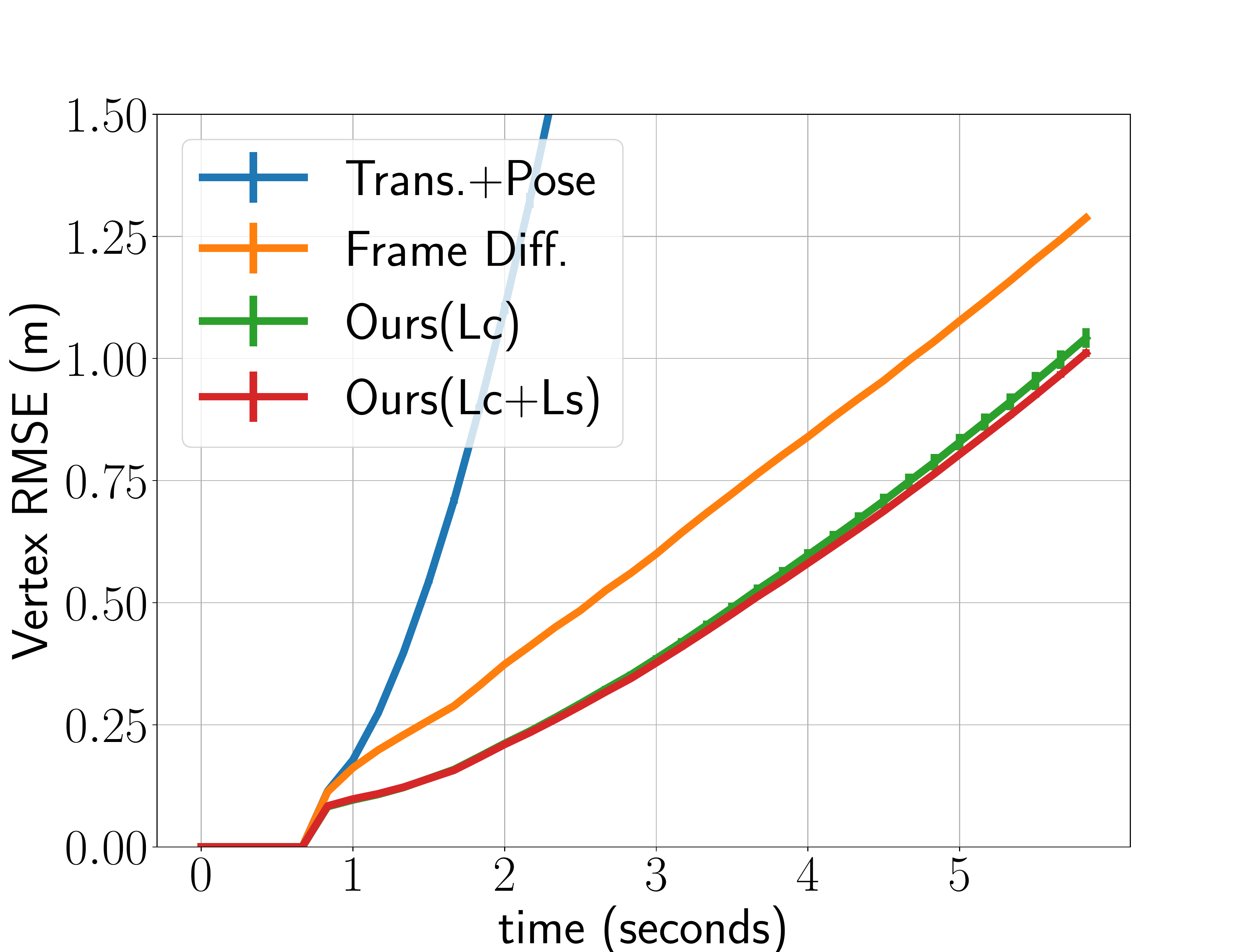}
\caption{ }
\label{fig:MTP_vertex_zoom}
\end{subfigure}
\begin{subfigure}{0.22\textwidth}
\includegraphics[width=\textwidth, trim={0 0mm 7mm 29mm},clip]{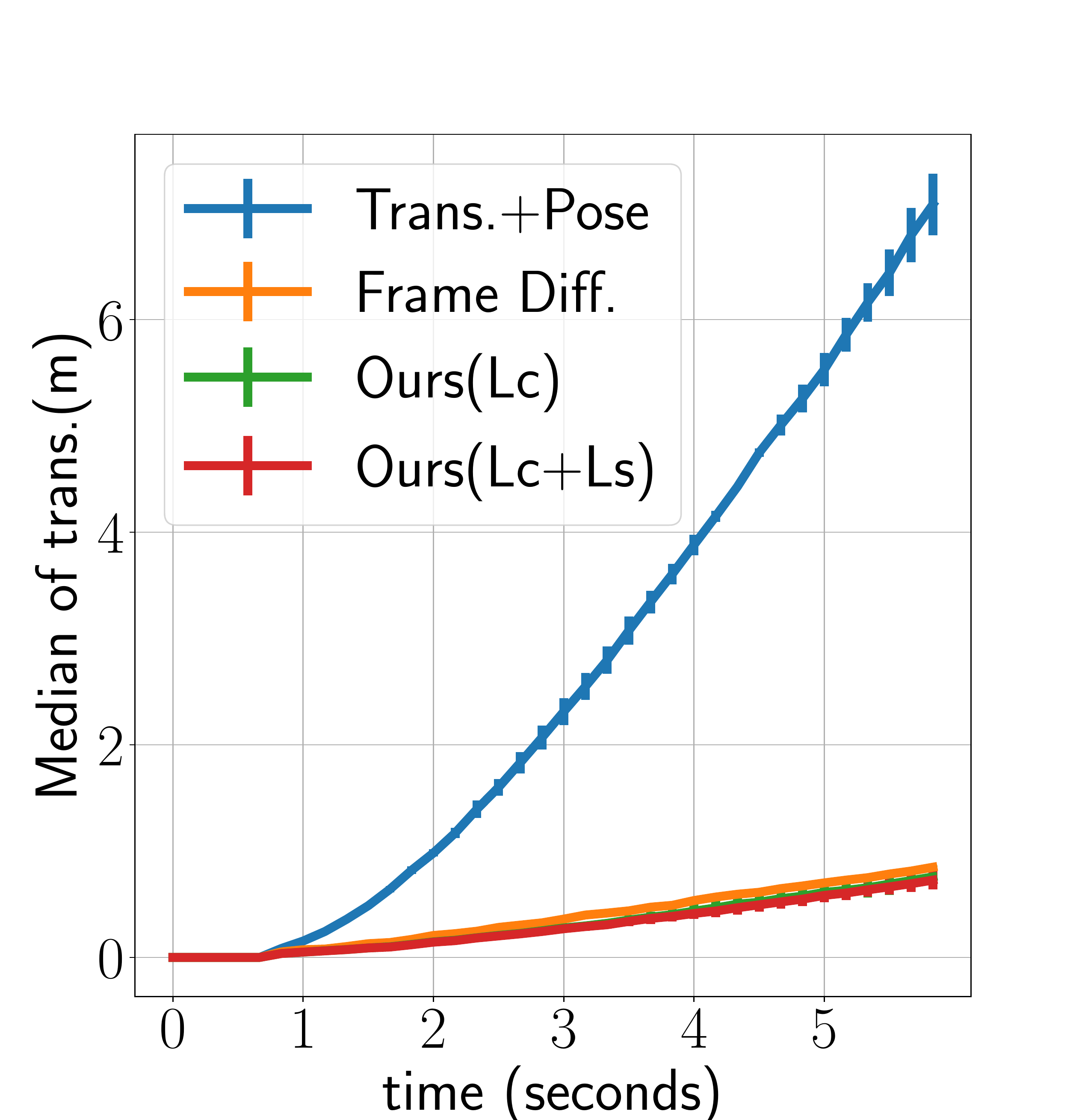}
\caption{ }
\label{fig:MTP_trans}
\end{subfigure}
\begin{subfigure}{0.26\textwidth}
\includegraphics[width=\textwidth, trim={0mm 0mm 0mm 29mm},clip]{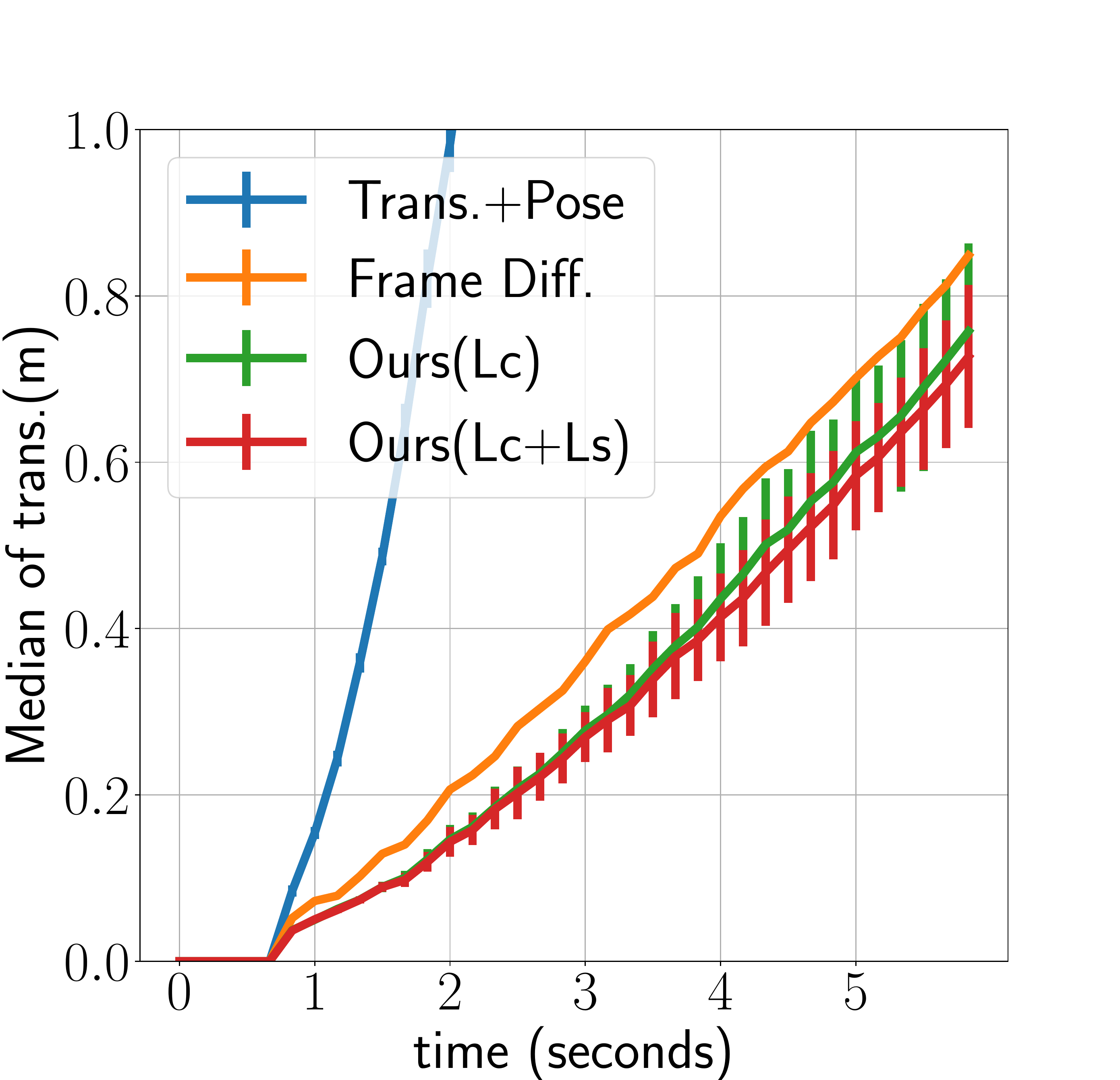}
\caption{ }
\label{fig:MTP_trans_zoom}
\end{subfigure}
\caption{\textcolor{black}{The plots of MTP results. X-axis: time in seconds(s) with an increment of 1/6s (time per frame). Y-axis: vertex RMSE and median of translation RMSE results with error bar. (a)(c) Overall MTP comparison results between our proposed method and baseline. (b)(d) Zoomed-in view.}}
\label{fig:MTP}
\end{figure}

\textcolor{black}{Given five observed frames, we ran multi-timestep prediction for 31 timesteps into the future to form a sequence of approximately six seconds in total. We evaluated on 196 in-the-wild pedestrian sequences with equal or longer than 36 frames in the PedX dataset.  Figure~\ref{fig:MTP_vertex} and~\ref{fig:MTP_vertex_zoom} show the vertex RMSE results for MTP prediction of our proposed network and all baseline methods, compared with the ground truth (observed) poses. Note the first five timesteps were given as training data, so the corresponding errors were zero across all methods. We observed that the errors of the comparison methods increased drastically over time, as the noise overcame the system. On the other hand, our proposed network was able to achieve much lower error in comparison.}

\begin{figure}[h]
\centering   
\includegraphics[width=0.5\textwidth, trim={0 0mm 0mm 0},clip]{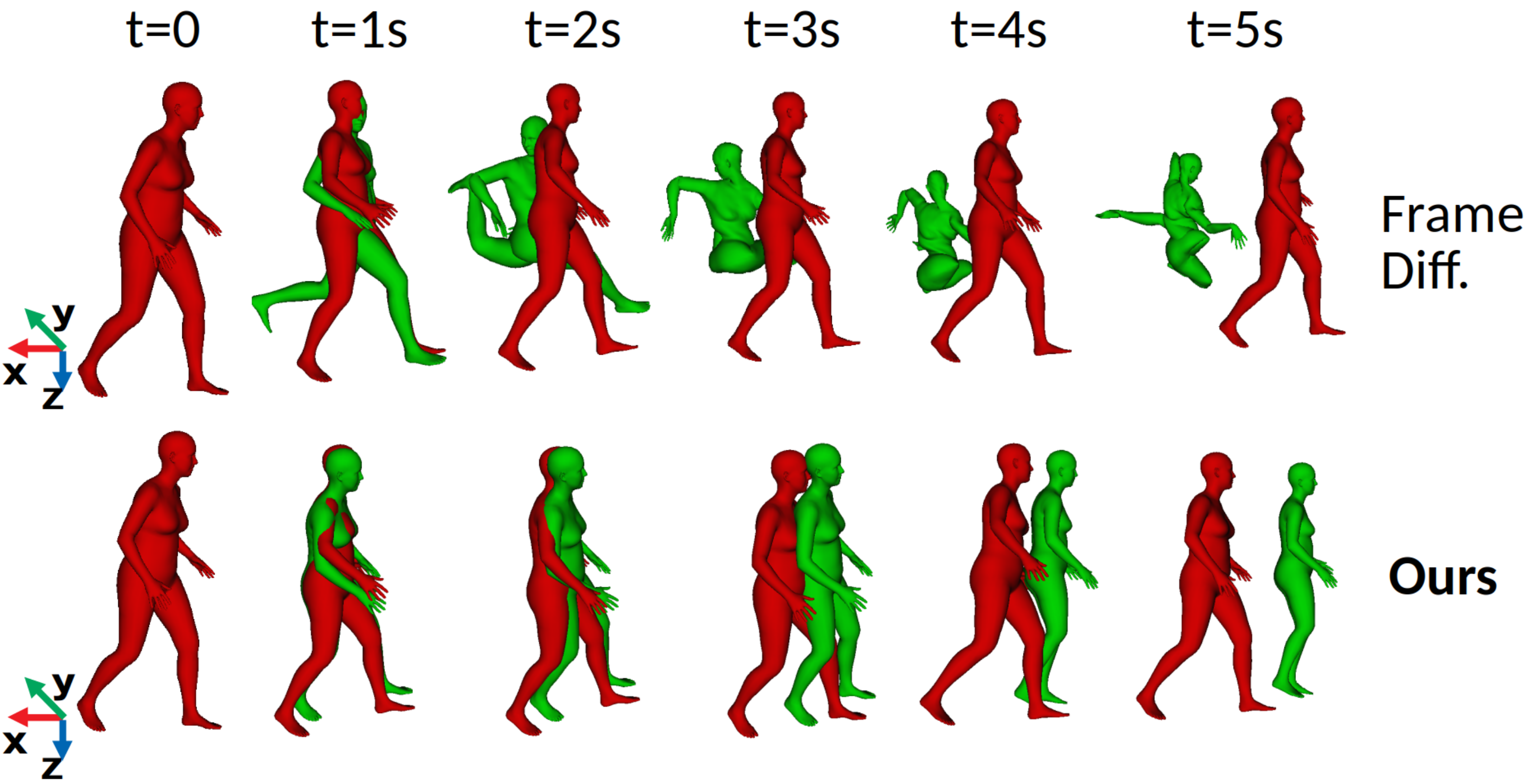}
\caption{\textcolor{black}{A qualitative example for MTP prediction. The green meshes are predicted poses and the red meshes are the ``ground truth'' labels as optimized in \cite{kim2018pedx}. Both methods have low translation error, but our method preserves a steady walking gait while frame difference method yields unrealistic and biomechanically infeasible poses.}}
\label{fig:MTP_compare_viz_2}
\end{figure}

\textcolor{black}{We further analyzed our MTP prediction performance. We noticed that, in several cases, pedestrians move with a high degree of stochasticity (sudden turns, crossing the crosswalk multiple times in different directions, etc.). In these cases, the network predicted the person walking with a smooth trajectory but going to a completely different direction, and the error ends up being really large ($\approx$5 meters after 6 seconds).  This effect of stochasticity in human motion was also reported in \cite{butepage2018anticipating, fragkiadaki2015recurrent, jain2016structural}. These cases contributed to a higher mean vertex RMSE, especially when predicting far into the future.  When we plot the median of translation error as shown in  Figure~\ref{fig:MTP_trans} and~\ref{fig:MTP_trans_zoom}, our network was able to achieve approximately 10cm error after one second and less than 80cm after 6 seconds, while comparison methods can be up to 7 meters off. The Frame Difference baseline method also did reasonably well in translation RMSE (still not as good as our method). However, if we look at the actual predicted pose of the person, the frame difference method yields rather unrealistic and biomechanically infeasible poses, likely due to the linear pose prediction based simply on frame difference. Our method, on the other hand, maintains a steady walking gait, as shown in Figure~\ref{fig:MTP_compare_viz_2}.\textsuperscript{\ref{footnote}}}


\textcolor{black}{Our proposed network was implemented in Python 3.6 using the Keras framework \cite{chollet2015keras}.} With the current unoptimized code, the prediction takes approximately 1ms for each person in each frame on a desktop computer with Intel i7 3.60GHz CPU with two NVIDIA TITAN X GPUs. Future work will include applying the approach in real-time data capture and prediction in autonomous vehicle applications.

\section{Conclusion}
\label{sec:conclusion}

This paper proposes bio-LSTM, a biomechanically inspired recurrent neural network for 3D pedestrian pose and gait prediction. Bio-LSTM is able to predict the global location and 3D full-body mesh with articulated body pose in the metric space. Evaluating our method on PedX, a large-scale, in-the-wild urban intersection pedestrian dataset, we predict more accurate and biologically feasible body poses than the current state-of-the art. Furthermore, our network is robust to noise in the training data.

Currently, this work is focused on pedestrian pose prediction at urban intersections, which has applications in planning human-oriented, pedestrian-friendly intersections and smart cities. \textcolor{black}{In addition, our work may benefit gait studies of bipedal robots and be applied to the monitoring and development of clinical gait rehabilitation systems. We provided detailed analysis on a variety of human actions in the intersection environment and showed improved prediction results on all pedestrian (non-cyclist) actions. It is possible to extend this work to predict other activities, such as running, as well.}  Also, we currently assume independence between the pedestrians.  It would be interesting to consider constraints to accommodate multiple persons in the same space \cite{zanfir2018monocular}. Future work will also include incorporating pedestrian-pedestrian and car-pedestrian interactions. 

Our novel objective function took the first step in imposing biomechanical constraints on pedestrian gait prediction. However, there are many aspects of human gait characteristics that can be further investigated, such as the dynamical asymmetry in gaits \cite{liu2006improved} and change of foot pressure on different parts of the foot in a human gait cycle \cite{kong2008smooth, winter1995human}. In addition, although the body shapes were optimized in the previous work and used in our prediction, we did not make a point to differ between genders and simply used a gender-neutral SMPL mesh. However, it has been shown in literature that men and women have different stride lengths and it is possible to distinguish individual gait for each person \cite{troje2005person}. By using the frame difference (the $L_c$ constraint), in a way, we inherently assume each person maintains their own stride and personal gait characteristics. However, it is possible to further investigate such individual gait characteristics in pose prediction.

In addition, it would be interesting to extend our current work to varying sequence length (varying $l$) and sequences with finer time resolution. \textcolor{black}{It is also possible to explore imposing biomechanical constraints on alternative network architectures such as the QuaterNet \cite{pavllo2018quaternet}.} Future work will also include combining pose estimation and prediction for an end-to-end pose analysis system. 


\section*{Acknowledgment}
The authors thank Wonhui Kim for making the 3D pose estimation results on the PedX data available \cite{kim2018pedx}. The authors also thank Charles Barto for his work in visualizing the 3D SMPL mesh models.

\bibliographystyle{IEEEtran} 
\bibliography{bibtex_entries}

\end{document}